\PassOptionsToPackage{authoryear,sort}{natbib}

\documentclass{article}

\usepackage{hyperref}
\usepackage{url}

\usepackage{pifont}
\usepackage[authoryear,round,sort]{natbib}
\setcitestyle{authoryear,round}

\usepackage{newtxtext}

\usepackage{booktabs}
\usepackage{colortbl}
\usepackage{xcolor}
\usepackage{multirow}
\usepackage{diagbox}
\usepackage{tablefootnote}

\usepackage{amsmath,amsthm,amsfonts,amssymb}
\usepackage{nicefrac}
\usepackage{bm}
\usepackage{dsfont}

\usepackage{microtype}
\usepackage{subcaption}
\usepackage[font=small]{caption}
\usepackage{graphicx}
\usepackage{wrapfig}
\usepackage{multicol}
\usepackage{placeins}

\usepackage[inline]{enumitem}
\usepackage[capitalise]{cleveref}
\usepackage{comment}
\usepackage{etoolbox}
\usepackage{listings}
\usepackage{tcolorbox}
\usepackage{xspace}

\usepackage{algorithm,algorithmicx,algpseudocode}

\usepackage{todonotes}

\newsavebox{\memtable}
\usepackage{makecell}

\usepackage{amsmath,amsfonts,bm}

\def\eqref#1{equation~\ref{#1}}

\def\1{\bm{1}}

\def\va{{\bm{a}}}
\def\vb{{\bm{b}}}

\def\ve{{\bm{e}}}

\def\vg{{\bm{g}}}

\def\vk{{\bm{k}}}

\def\vm{{\bm{m}}}

\def\vo{{\bm{o}}}
\def\vp{{\bm{p}}}
\def\vq{{\bm{q}}}
\def\vr{{\bm{r}}}

\def\vu{{\bm{u}}}
\def\vv{{\bm{v}}}
\def\vw{{\bm{w}}}
\def\vx{{\bm{x}}}

\def\mA{{\bm{A}}}
\def\mB{{\bm{B}}}
\def\mC{{\bm{C}}}
\def\mD{{\bm{D}}}

\def\mG{{\bm{G}}}

\def\mI{{\bm{I}}}

\def\mK{{\bm{K}}}

\def\mM{{\bm{M}}}

\def\mO{{\bm{O}}}
\def\mP{{\bm{P}}}
\def\mQ{{\bm{Q}}}

\def\mS{{\bm{S}}}

\def\mU{{\bm{U}}}
\def\mV{{\bm{V}}}
\def\mW{{\bm{W}}}

\DeclareMathAlphabet{\mathsfit}{\encodingdefault}{\sfdefault}{m}{sl}
\SetMathAlphabet{\mathsfit}{bold}{\encodingdefault}{\sfdefault}{bx}{n}

\newcommand{\R}{\mathbb{R}}

\newcommand{\softmax}{\mathrm{softmax}}

\definecolor{tabblue}{HTML}{1F77B4}
    \definecolor{taborange}{HTML}{FF7F0E}
    \definecolor{tabred}{HTML}{D62728}
    \definecolor{tabcyan}{HTML}{17BECF}
    \definecolor{tabgreen}{HTML}{2CA02C}
    \definecolor{Green}{HTML}{2CA02C}
    \definecolor{darkgreencite}{rgb}{0.0,0.5,0.0}
    \definecolor{blue_nice}{HTML}{dbe9f7}
    \definecolor{darkgreen}{HTML}{d9f2d2}
    \definecolor{rmmblue}{HTML}{dbeaf7}
    \definecolor{dg}{RGB}{0,102,51}
    
\pgfdeclareverticalshading{greenBlueRed}{100bp}{color(0bp)=(red!30!white);
      color(35bp)=(red!30!white);
      color(46bp)=(rmmblue);
      color(55bp)=(rmmblue);
      color(62bp)=(darkgreen);
      color(100bp)=(darkgreen)
    }
    
\hypersetup{
  colorlinks=true,
  citecolor=green!40!black, 
  linkcolor=blue,
  urlcolor=magenta
}
\newcolumntype{a}{>{\columncolor{green!5}}l}

\theoremstyle{definition}   

\newtheorem{theorem}{Theorem}

\newtheorem{proposition}[theorem]{Proposition}
\newtheorem{definition}{Definition}

\AddToHook{env/theorem/begin}{\crefalias{section}{theorem}}
\AddToHook{env/lemma/begin}{\crefalias{section}{lemma}}
\AddToHook{env/corollary/begin}{\crefalias{section}{corollary}}
\AddToHook{env/observation/begin}{\crefalias{section}{observation}}
\AddToHook{env/proposition/begin}{\crefalias{section}{proposition}}
\AddToHook{env/definition/begin}{\crefalias{section}{definition}}
\AddToHook{env/remark/begin}{\crefalias{section}{remark}}
\AddToHook{env/claim/begin}{\crefalias{section}{claim}}

\newcommand{\xmark}{\ding{55}}

\newcommand*\colourcheck[1]{\expandafter\newcommand\csname #1check\endcsname{\textcolor{green!80!black}{\ding{52}}}}

\newcommand*\colourxmark[2]{\expandafter\newcommand\csname #2check\endcsname{\textcolor{#2}{\ding{5}}}}

\colourcheck{green}
\newcommand{\githubrepo}[2]{\href{#2}{\includegraphics[height=1.2em]{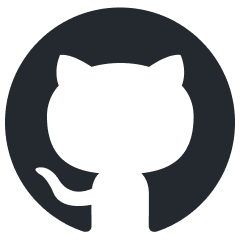}\hspace{0.3em}\texttt{#1}}}

\newcommand{\blogrepo}[2]{\href{#2}{\includegraphics[height=1.2em]{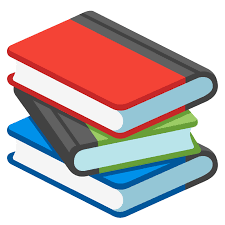}\hspace{0.3em}\texttt{#1}}}

\newtoggle{comments}
\toggletrue{comments}

\iftoggle{comments}{
  \newcommand{\AG}[1]{\textcolor{magenta}{[AG: #1]}}
  \newcommand{\AB}[1]{\textcolor{blue}{[AB: #1]}}
  \newcommand{\AAF}[1]{\textcolor{brown}{[AA: #1]}}
}{
  \newcommand{\AG}[1]{}
  \newcommand{\AB}[1]{}
  \newcommand{\AAF}[1]{}
}

\NewDocumentCommand{\CITE}{o}{\IfNoValueTF{#1}
    {\textcolor{red}{[CITE]}}
    {\textcolor{red}{[#1]}}}

\newcommand{\ours}{Raven\xspace}

\makeatletter
\newcommand{\iftwocolumn}[2]{\if@twocolumn #1\else #2\fi}
\makeatother

\newtcolorbox{greenbox}[1][]{colback=green!4,
  colframe=black,
  boxrule=0.3pt,
  boxsep=-4pt,
  #1}
\newtcolorbox{whitebox}[1][]{colback=white!4,
  colframe=black,
  boxrule=0.3pt,
  boxsep=0.1pt,
  left=0.1pt,
  right=2pt,
  top=0pt,
  bottom=0pt,
  #1}
\newtcolorbox[auto counter, number within=section]{gbox}[2][]{colback=green!10!white,
  colframe=green!80!black,
  boxrule=0.8mm,
  arc=4mm,
  width=0.2mm,
  #1}

\newcommand{\hred}[2][red!20]{\mathchoice {\colorbox{#1}{$\displaystyle#2$}}{\colorbox{#1}{$\textstyle#2$}}{\colorbox{#1}{$\scriptstyle#2$}}{\colorbox{#1}{$\scriptscriptstyle#2$}}}

\newcommand{\hgray}[2][gray!20]{\mathchoice {\colorbox{#1}{$\displaystyle#2$}}{\colorbox{#1}{$\textstyle#2$}}{\colorbox{#1}{$\scriptstyle#2$}}{\colorbox{#1}{$\scriptscriptstyle#2$}}}

\newcommand{\hgreen}[2][darkgreen]{\mathchoice {\colorbox{#1}{$\displaystyle#2$}}{\colorbox{#1}{$\textstyle#2$}}{\colorbox{#1}{$\scriptstyle#2$}}{\colorbox{#1}{$\scriptscriptstyle#2$}}}

\newcommand{\hblue}[2][blue_nice]{\mathchoice {\colorbox{#1}{$\displaystyle#2$}}{\colorbox{#1}{$\textstyle#2$}}{\colorbox{#1}{$\scriptstyle#2$}}{\colorbox{#1}{$\scriptscriptstyle#2$}}}

\newlength{\defbaselineskip}
\usepackage{authblk}

\title{\ours: High-Recall Sequence Modeling with Sparse Memory Routing}
\date{}
\author{
\makebox[\textwidth][c]{\begin{minipage}[t]{0.45\textwidth}
    \centering
    \textbf{Arshia Afzal}\thanks{Equal contribution (alphabetical order)}\\
    \vspace{-3mm}
    EPFL\\
    \texttt{arshia.afzal@epfl.ch}
\end{minipage}
\hspace{-16mm}
\begin{minipage}[t]{0.45\textwidth}
    \centering
    \textbf{Aviv Bick}\textsuperscript{*}\\
    \vspace{-3mm}
    Carnegie Mellon University\\
    \texttt{abick@cs.cmu.edu}
\end{minipage}}
\par\vspace{1mm}
\makebox[\textwidth][c]{\begin{minipage}[t]{0.32\textwidth}
    \centering
    \textbf{Eric P. Xing}\\ 
    \vspace{-3mm}
    Carnegie Mellon University,\\
    MBZUAI
\end{minipage}\hfill
\begin{minipage}[t]{0.32\textwidth}
    \centering
    \textbf{Volkan Cevher}\\
    \vspace{-3mm}
    EPFL
\end{minipage}\hfill
\begin{minipage}[t]{0.32\textwidth}
    \centering
    \textbf{Albert Gu}\\
    \vspace{-3mm}
    Carnegie Mellon University,\\
    Cartesia AI
\end{minipage}}
}
\begin{document}
\maketitle

\begin{abstract}
\noindent
Long-context recall in linear-time sequence models highlights a tradeoff in how they write to memory.
State-based linear models, such as state-space models (SSMs) and linear Transformers, write \emph{densely}, updating the entire state for each newly arrived token, which leads to interference and makes specific past tokens hard to recover.
Sliding-window attention (SWA) exhibits the opposite behavior: it writes \emph{sparsely} by storing explicit token representations, but only within a fixed window, so recall drops once the relevant token is evicted.
Interpolating between these models, we introduce \textit{\textbf{\ours}}, a linear-time sequence model that maintains a fixed set of memory slots and, at each step, decays and updates only a selected subset via learned, \textit{input-dependent routing}. This lets \ours mitigate SWA's position-based overwriting and hard eviction while reducing interference from dense state updates in SSMs, thereby preserving long-range content much more effectively.
Across recall-intensive benchmarks, \ours is competitive with or outperforms prior linear-time baselines, achieving strong long-context recall where both SWA and SSMs sharply degrade. It remains effective when extrapolating to context lengths as large as 16$\times$ its training length, with similar gains in hybrid architectures.
\end{abstract}

\begin{center}
    \githubrepo{Raven Code}{https://github.com/goombalab/raven} 
    \hspace{2.5mm} \hspace{2.5mm}
    \blogrepo{Raven Blog}{https://arshiaafzal.github.io/blog/2026/raven-part1/}
     \hspace{2.5mm} 
\end{center}

\section{Introduction}

Fixed-size memory models often match Transformers in language modeling but still fall short in long-range recall \citep{bick2504understanding}, despite being able to carry forward task-relevant features over arbitrarily long contexts.
This gap reflects how linear models manage history. 
State Space Models (SSMs) \citep{mamba,mamba2} and linear Transformer variants \citep{deltanet,yang2023gated,trans_rnn} are highly persistent---their state can, in principle, carry information \textit{indefinitely}---but they write \emph{densely}, updating the entire memory at every step, which makes individual items hard to preserve without interference. 
Sliding-window attention (SWA) exhibits the opposite tradeoff \citep{longformer}: it writes \emph{sparsely} by retaining explicit token representations only within a fixed recent window, enabling reliable in-window retrieval but \emph{no persistence} once older tokens are dropped. Together, these complementary limitations motivate separating \emph{where} information is written from \emph{how long} it persists in memory-based sequence models.
 
\begin{figure*}[t]
\centering
\includegraphics[width=0.9\textwidth]{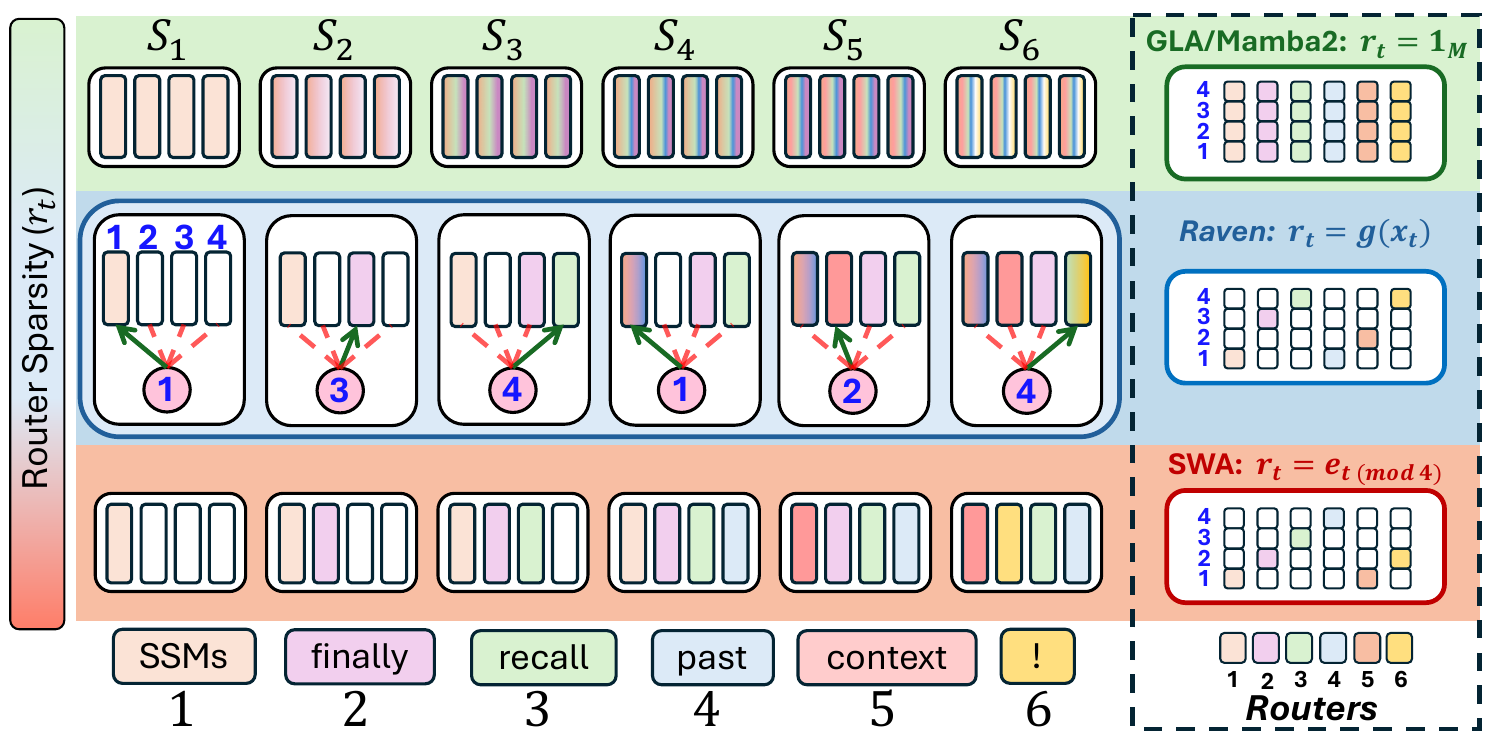}
\caption{
\textbf{\ours Overview.}
Visualization of three different sequence mixers using \textbf{\textit{Routing Slot Memories}}, with different router choices.  \textit{\textbf{(a)}} $\hred{\text{\textcolor{red!80!black}{SWA}}}$ memory allocation as \textit{first-in-first-out} strategy using a \textbf{one-hot vector $\ve_t$} as router. \textit{\textbf{(b)}} $\hgreen{\text{\textcolor{green!10!black}{SSM}}}$ memory allocation, which projects each token in all memory slots using a \textbf{dense, all-ones router $\mathbf{1}_M$} \textit{\textbf{(c)}} $\hblue{\text{\textbf{\ours}}}$ memory allocation, which uses a \textbf{selective router} for writes. Visualization uses a sequence of $T=6$ token and $M=4$ memory slots for the hidden state $\mS_t$ and $\text{Top}_K=1$ for \ours router.
}
\label{fig:phoenix}
\end{figure*}

 Motivated by this view, we propose \textbf{\textit{Routing Slot Memories (RSMs)}}, 
a class of sequence models that maintain a fixed set of memory slots and, at each step, (i) route newly written content into one or more of these slots and (ii) apply an explicit decay (forgetting) operator to the updated slot.
Under this view, SSMs and SWA emerge as two extremes on the routing axis: SSMs write densely with gradual forgetting, while SWA writes sparsely and deterministically with hard eviction --- making explicit a tradeoff that existing architectures have left implicit, as illustrated in \Cref{fig:phoenix}.

Designed to sit between these endpoints, we introduce \textbf{\ours}, an instantiation of RSMs that uses \textit{sparse input-dependent routing to update a selected subset of memory slots, and applies decay only to those updated slots}.
This lets the model better organize the memory, keeping specific tokens recoverable even when the context grows far beyond the training length.
Importantly, the \ours block simplifies prior linear models and 
\textit{does not rely on convolutional or sliding-window attention layers} for effective retrieval \citep{mamba,yang2023gated}. 

As instantiations of RSMs, \ours can be contrasted with both SWA and SSMs:

\textbf{Compared to SWA}, \ours selectively routes writes to memory and gradually decays existing content in the chosen slots, generalizing SWA's fixed window approach. This enables \ours to achieve perfect recall on Needle-in-a-Haystack benchmarks where SWA performs near zero, with substantial improvements on recall-heavy tasks in both standalone and hybrid settings across different model scales.

\textbf{Compared to SSMs}, \ours avoids dense full-state updates by restricting both writing and forgetting to a selected subset of memory slots, providing long-term persistence that prior SSMs lack. 
\ours is competitive with or outperforms recent SSMs and linear models---including Mamba-2, Gated Delta-Net (GDN), and Gated Linear Attention (GLA)---on recall-intensive benchmarks, while maintaining strong accuracy even when extrapolating 16$\times$ beyond those seen during training.

\section{Background} 
\label{sec:background}

\subsection{State-Space Models}

Linear state-based sequence models maintain an explicit, fixed-size memory that is updated sequentially.
Given an input $\vx=(\vx_1,\ldots,\vx_T)$, the model maintains a 2-dimensional memory matrix $\mS_t\in\mathbb{R}^{M\times d}$ for each head. Here, $M$ corresponds to the value head dimension, while $d$ corresponds to the query/key head dimension (aka the state expansion).
Because each of the $M$ rows of $\mS_t$ is updated independently in all models we consider, $M$ doubles as the number of \emph{memory slots}; \Cref{sec:RSM} adopts this reading throughout.
At step \(t\), the token \(\vx_t\in\mathbb{R}^{d_{\mathrm{model}}}\) is linearly projected to
\(\vq_t,\vk_t\in\mathbb{R}^d\) and \(\vv_t\in\mathbb{R}^M\)
\footnote{
SSMs are often written with ($\vx,\mB,\mC$);
we use \((\vv,\vk,\vq)\) for a read/write interpretation and to stay consistent with later attention-based special cases (e.g., sliding-window attention).
},
and updates memory via
\begin{equation}
\label{ssm_equation}
\underbrace{\mS_t}_{\text{Memory}}
=
\underbrace{\mS_{t-1}\mA_t}_{\text{Decay}}
\;+\;
\underbrace{\vv_t \vk_t^\top}_{\text{Write}},
\qquad
\underbrace{\vo_t}_{\text{Output}}
=
\underbrace{\mS_t \vq_t}_{\text{Read}}.
\end{equation}
Here, \(\mA_t\in\mathbb{R}^{d\times d}\) is a (possibly input-dependent) \textit{channel-wise} decay commonly known as forget gate \citep{yang2023gated}.
Intuitively, \(\mA_t\) controls how past memory is retained, \(\vv_t\vk_t^\top\) writes new content, and \(\vo_t=\mS_t\vq_t\) reads from memory.
Some variants apply a feature map \(\phi(\cdot)\) to these projections \citep{trans_rnn}; for simplicity, we absorb it into \(\vq_t,\vk_t,\vv_t\).

SSM variants mainly differ in the structure of \(\mA_t\) (see Table~2 in \citet{yang2023gated}).
For example, GLA uses a diagonal gate \citep{yang2023gated}, Mamba-2 a scalar gate \citep{mamba2}, and DeltaNet a delta-rule update \(\mA_t = I - \vk_t\vk_t^\top\) \citep{deltanet,schlag2021linear}.
In the diagonal case \(\mA_t=\mathrm{diag}(\va_t)\), \Cref{ssm_equation} becomes
\begin{align}
\label{ssm_hadamard}
\mS_t = \mS_{t-1} \odot \va_t + \vv_t \vk_t^\top,
\qquad
\vo_t = \mS_t \vq_t.
\end{align}
where $\odot$ is columnwise Hadamard product (broadcast over the $M$ rows).

\subsection{Sliding-Window Attention as a Dual-State SSM}

Sliding-Window Attention (SWA) maintains a key/value cache for the most recent $M$ tokens and restricts attention scores to this window.
Unlike SSMs, which compress all history into a fixed-size state, SWA stores token-level memories for the last $M$ steps.
Nevertheless, SWA admits a state-space view: it maintains two coupled states (key and value caches) updated by a linear recurrence.

More precisely, let $\mS_t^{k}, \mS_t^{v} \in \mathbb{R}^{M \times d}$ denote the key and value caches at time $t$, where each row corresponds to a memory slot.
SWA updates the cache using a \textit{First-In-First-Out} (FIFO) ring buffer.
Let
\[
m_t = 1 + ((t-1)\bmod M),
\]
denote the slot index in the FIFO buffer, and let $\ve_t\in\mathbb{R}^{M}$ be the one-hot vector with $\ve_t[m_t]=1$.
Then SWA overwrites exactly one slot per step:
\begin{align}
\mS^{k}_t = (\mI - \ve_t\ve_t^\top)\mS^{k}_{t-1}  + \ve_t\vk_t^\top,
\iftwocolumn{\\}{\qquad}
\mS^{v}_t = (\mI - \ve_t\ve_t^\top) \mS^{v}_{t-1} + \ve_t\vv_t^\top.
\end{align}
Since $\ve_t$ is one-hot, this simplifies to
\begin{align}
\mS^{k}_t =  (\mathbf{1}_M-\ve_t) \odot \mS^{k}_{t-1}  + \ve_t\vk_t^\top,
\iftwocolumn{\\}{\qquad}
\mS^{v}_t = (\mathbf{1}_M-\ve_t) \odot\mS^{v}_{t-1} + \ve_t\vv_t^\top.
\label{eq:swa_recurrence}
\end{align}
Thus, SWA computes attention over the window:
\begin{align}
\vo_t= (\mS^v_t)^\top \,\softmax\!\left(\mS^k_t \vq_t\right),
\label{swa_readout}
\end{align}
where the softmax is over the $M$ slots.
For $t<M$, one can either initialize $\mS_0^{k,v}=0$ and apply a validity mask in \eqref{swa_readout}, or attend only over the filled prefix.
From \Cref{eq:swa_recurrence}, SWA applies a binary \textit{slot-wise} decay over the $M$ slots and removes the earliest token from memory.
Stacking the two states along the feature dimension makes this connection more explicit:
\[
\mS_t = \begin{bmatrix} \mS^k_t & \mS^v_t \end{bmatrix} \in \mathbb{R}^{M \times 2d},
\qquad
\vu_t = \begin{bmatrix} \vk_t \\ \vv_t \end{bmatrix} \in \mathbb{R}^{2d},
\]
which updates as
\begin{equation}
\mS_t = (\mathbf{1}_{M} - \ve_t) \odot \mS_{t-1} + \ve_t \vu_t^\top,
\end{equation}
casting SWA exactly as an SSM with state size $M \times 2d$.

\section{Routing Slot Memories}
\label{sec:RSM}

As shown in \Cref{sec:background}, linear sequence models such as SSMs and SWA both maintain a finite-dimensional state $\mS_t \in \R^{M \times d}$ that compresses the entire history into $M$ memory slots.

The central question is therefore not whether to forget---as all finite-state models inevitably must---but rather \textit{where} information should be written and \textit{which slots} should be preserved. We formalize this perspective through the lens of \textbf{Routing Slot Memories (RSMs)}, a unified class of slot-separable linear recurrences that makes the write location explicit via a \textit{routing vector}.

\subsection{Memory Slots and Routing}

We organize the hidden state as a matrix $\mS_t \in \R^{M \times d}$, where each row $\mS_t[i]$ is an independent memory \textit{slot}.
The general recurrence and readout are:
\begin{equation}
\mS_t = g_t(\mS_{t-1}, \vx_t), \qquad \vo_t = f(\mS_t, \vx_t).
\label{eq:update_n_readout}
\end{equation}
The key structural property we require is \textbf{slot separability}: each slot updates independently of all others.

\begin{definition}[Slot-Separable Update]
A state update is \emph{slot-separable} if for every $i$:
\begin{equation}
\mS_t[i] = g_{t,i}\!\bigl(\mS_{t-1}[i],\, \vx_t\bigr).
\label{eq:slot-separable-def}
\end{equation}
\end{definition}

For linear recurrences, slot separability has a clean algebraic characterization.

\begin{proposition}[Slot Separability of a Linear Update]
\label{cor:linear-slot-sep}
Consider the linear update
\begin{equation}
\mS_t \;=\; \mD_t \mS_{t-1}\mA_t \;+\; \mU_t,
\label{eq:bilinear-recurrence}
\end{equation}
where $\mU_t$ is independent of $\mS_{t-1}$. Then
\begin{equation}
\text{Update is slot-separable}
\quad\Longleftrightarrow\quad
\mD_t \text{ is diagonal.}
\label{eq:linear-slot-sep-crux}
\end{equation}
Diagonality of $\mD_t$ makes the transition row-wise, yielding $M$ independent slots \emph{(see \Cref{slotsepdt})}.
\end{proposition}

\Cref{eq:bilinear-recurrence}, which incorporates decays on both sides of the linear update, was also introduced by \citet{zhong2025understanding}. Slot separability constrains how each slot is updated, but does not specify where new information is written. To make the write location explicit, we introduce a router.

\begin{definition}[Router]
A \emph{router} $\vr_t = r(\vx_t, t) \in \R^M$ is a vector that determines, at each time step, how new information is distributed across slots. The entry $\vr_t[i]$ specifies the write intensity for slot $i$, gating how much of the incoming information is written to that slot. Routers may depend on the current input $\vx_t$ (input-dependent routing) or only on time $t$ (time-dependent routing).
\end{definition}

\subsection{The Spectrum of Routing Slot Memories}

\begin{definition}[RSM]
A \emph{Routing Slot Memory} is a slot-separable linear update of the form:
\begin{equation}
\mS_t
= \underbrace{(\mathbf{1} - \vr_t) \odot \mS_{t-1}}_{\text{preserved memory}}
+ \underbrace{\vr_t \odot \bigl(\mD_t\,\mS_{t-1}\,\mA_t + \mU_t\bigr)}_{\text{updated memory}},
\label{eq:RSM}
\end{equation}
where $\odot$ broadcasts row-wise, $\mD_t \in \R^{M \times M}$ is diagonal, $\mA_t \in \R^{d \times d}$ mixes features, $\mU_t \in \R^{M \times d}$ is the write content derived from $\vx_t$, and $\vr_t \in \R^M$ is the routing vector.
\end{definition}

The semantics are clean: when $\vr_t[i] = 0$, slot $i$ is frozen---its content persists exactly. When $\vr_t[i] = 1$, it is fully updated. Intermediate values interpolate. A slot written at time $j$ retains its content until the router selects it again, enabling long-lived storage with targeted updates.

\subsection{Dual-State Models as RSMs}
\label{subsec:dual_state_RSM}

Dual-state models maintain separate key and value memories, $\mS^k_t, \mS^v_t \in \R^{M \times d}$, and differ from standard SSMs in that they apply \textit{slot-wise} gating via $\vr_t$ rather than channel-wise decay. The two main instantiations differ in a single design choice: whether the router is input-dependent.

\paragraph{SWA.}
Sliding Window Attention~\citep{longformer} is a maximally sparse RSM. Its KV-cache can be written as the recurrence:
\begin{equation}
\mS^k_t = (\mathbf{1}_M - \ve_t) \odot \mS^k_{t-1} + \ve_t\, \vk_t^\top,
\qquad
\mS^v_t = (\mathbf{1}_M - \ve_t) \odot \mS^v_{t-1} + \ve_t\, \vv_t^\top,
\label{eq:swa_RSM}
\end{equation}
where $\ve_t$ is the $(t \bmod M)$-th basis vector. The router is time-dependent and one-hot: exactly one slot is overwritten per step, with no decay:
\[
\mA_t = \mI, \quad \mD_t = \mathbf{0}, \quad \vr_t = \ve_t, \quad \mU_t=[\vk_t,\vv_t]
\]
The SWA router performs hard deletion, since the oldest slot is unconditionally erased. The write content is the stacked key and value vector $[\vk_t,\vv_t]\in\R^{2d}$.

\paragraph{ABC.}
Attention with Bounded Memory Control~\citep{abc} uses dense input-dependent routing without any decay:
\begin{equation}
\mS^k_t = \mS^k_{t-1} + \vr_t\,\vk_t^\top,
\qquad
\mS^v_t = \mS^v_{t-1} + \vr_t\,\vv_t^\top,
\label{eq:abc_RSM}
\end{equation}
using a softmax router of $\vr_t=\softmax(\mW\vx_t)$. This way, the effect of each token across the $M$ memory slots is bounded and sums to 1. However, ABC lacks decay of historical information, so it is equivalent to an RSM with the following parameters:
\[
\mA_t = \mI, \quad \mD_t = \mI, \quad \vr_t = \softmax(\mW\vx_t), \quad \mU_t=[\vk_t,\vv_t]
\]

\paragraph{GSA.}
Gated Slot Attention~\citep{gsa} replaces SWA's fixed cyclic router with an input-dependent one:
\begin{equation}
\mS^k_t = (\mathbf{1} - \vr_t) \odot \mS^k_{t-1} + \vr_t\,\vk_t^\top,
\qquad
\mS^v_t = (\mathbf{1} - \vr_t) \odot \mS^v_{t-1} + \vr_t\,\vv_t^\top,
\label{eq:gsa_RSM}
\end{equation}
with $\vr_t = \sigma(\mW\vx_t)^{1/\tau}$. Each token now decides how strongly to update each slot, allowing the model to be more conservative about overwriting. However, the router is typically dense: every slot receives some write signal at every step, which limits the model's ability to isolate and protect specific memories. GSA in the RSM view has the parameters
\[
\mA_t = \mI, \quad \mD_t = \mathbf{0}, \quad \vr_t = \sigma(\mW\vx_t)^{1/\tau}, \quad \mU_t=[\vk_t,\vv_t]
\]
with $\tau$ being a temperature parameter for the decay. One of the main differences between GSA and ABC in the RSM view is the parameter $\mD_t$, which removes state decay in ABC.

\subsection{State-Space Models as RSMs}
\label{subsec:ssm_RSM}

Standard diagonal SSMs are single-state models that are slot-separable and correspond to \textbf{dense routing}: $\vr_t = \mathbf{1}_M$ at every step. Every slot is updated unconditionally; the only mechanism for selective retention is decay. Mamba-1~\citep{mamba}, Mamba-2~\citep{mamba2}, and GLA~\citep{yang2023gated} follow this template, differing only in the granularity of their decay, as they all use a dense all-ones router:
\begin{align}
\text{Mamba-2:} \quad & \mS_t[i] = a_t\,\mS_{t-1}[i] + \vv_t[i]\,\vk_t^\top
    && \mD_t=\mI, \quad \mA_t = a_t && \text{(\textit{shared scalar} decay)},\\
\text{GLA/Mamba-1:} \quad & \mS_t[i] = \mS_{t-1}[i] \odot \va_t + \vv_t[i]\,\vk_t^\top
    && \mD_t=\mI, \quad \mA_t = \text{diag}(\va_t)
    && \text{(\textit{channel-wise} decay)}.
\end{align}
One can also transpose the hidden state $\mS_t^\top$ and apply the multiplications from the opposite side, resulting in an equivalent formulation that matches the notation used in other models.

Delta Networks~\citep{deltanet} also use dense routing ($\vr_t = \mathbf{1}_M$) but replace scalar or diagonal decay with a rank-one forget gate:
\begin{equation}
\mS_t = \mS_{t-1}(\mI - \beta_t\,\vk_t\vk_t^\top) + \beta_t\,\vv_t\vk_t^\top.
\end{equation}
Since the transition acts by right-multiplication ($\mD_t = \mI$), rows do not mix and slots remain separable. \Cref{tab:models} summarizes how all models fit within the RSM framework.

\paragraph{The routing/forgetting tradeoff.}
The RSM view exposes a fundamental tension across existing architectures. SSMs write densely to every slot and delegate all forgetting to decay, which is effective for smooth compression but unable to protect any slot from interference. SWA writes sparsely but is input-blind, discarding information by position rather than content. GSA and ABC~\citep{abc} gain input-dependent routing, but their dense writes still expose every slot at every step. The missing combination---\textbf{sparse}, \textbf{input-dependent} routing paired with \textbf{explicit decay}---is precisely what \ours introduces in \Cref{sec:ours}.

\begin{table*}
\centering
\small
\renewcommand{\arraystretch}{0.3}
\caption{
\textbf{Routing Slot Memories.} Design of the \textit{Router} and \textit{Decay} components across multiple architectures. The bar on the right along with row colors display each model's spectrum as a function of router sparsity, with $\hred{\text{\textcolor{red!80!black}{SWA}}}$ and $\hgreen{\text{\textcolor{green!10!black}{SSM}}}$ marking the two extremes. For SSMs, the content matrix ($\vu_t$) is \textbf{\textit{low-rank}}, whereas SWA-like models use a \textbf{\textit{sparse}} stacking of $\vk_t$ and $\vv_t$. $f(\cdot)$ denotes the softmax.  
}
\resizebox{\textwidth}{!}{\begin{tabular}{p{4.5cm} | p{2.7cm} p{1.9cm} p{2.8cm} p{2cm}} 
\toprule
\textbf{Model} &
\textbf{Router ($\vr_t$)} &
\textbf{Content ($\mU_t$)} &
\textbf{Decay ($\mA_t$)} &
\textbf{Readout ($\vo_t $)} \\
\midrule
\rowcolor{darkgreen}
LinAtt \citep{trans_rnn} & 
$\mathbf{1}_M$  & $\vv_t\vk_t^\top$ & $\mI$ & $\mS_t\vq_t$ \\
\rowcolor{darkgreen!60}
RetNet \citep{retnet} &  $\mathbf{1}_M$ &  $\vv_t\vk_t^\top$ &
$\gamma$ & $\mS_t\vq_t$ \\
\rowcolor{darkgreen!60}
GLA \citep{yang2023gated} & $\mathbf{1}_M$ & $\vv_t\vk_t^\top$ & $\text{diag}\left(\sigma(\mW\vx_t)\right)^{1/\tau}$ & $\mS_t\vq_t$ \\ 
\rowcolor{darkgreen!60}
Mamba-2 \citep{mamba2} & $\mathbf{1}_M$ & $\vv_t\vk_t^\top$ & $a_t$ & 
$\mS_t\vq_t$ \\
\rowcolor{darkgreen!60}
GDN \citep{yang2024gated} & $\mathbf{1}_M$ & $\vv_t\vk_t^\top$ & $a_t(\mI-\vk_t\vk_t^\top)$ & 
$\mS_t\vq_t$ \\
\rowcolor{rmmblue} 
\textbf{\ours }   & 
$
{\vg_t}/{\mathbf{1}^\top \vg_t}
$ & $[\vk_t \hspace{1mm} \vv_t]^\top$ & $\mI$
& $(\mS^v_t)^\top \,f\!\left(\mS^k_t\vq_t\right)$ \\
\rowcolor{red!5!white}
GSA \citep{gsa} &  $\mathbf{1}_M - \sigma(\mW\vx_t)^{1/\tau}$  & $[\vk_t \hspace{1mm} \vv_t]^\top$ 
& $\mI$
& $(\mS^v_t)^\top \,f\!\left(\mS^k_t\vq_t\right)$ \\
\rowcolor{red!5!white}
ABC \citep{abc} & 
$\softmax(\mW\vx_t)$ 
 & $[\vk_t \hspace{1mm} \vv_t]^\top$ & $\mI$  & $(\mS^v_t)^\top \,f\!\left(\mS^k_t\vq_t\right)$ \\
\rowcolor{red!10!white}
SWA \citep{longformer} & 
$\ve_t$ & $[\vk_t \hspace{1mm} \vv_t]^\top$ 
& $\mI$ & $(\mS^v_t)^\top \,f\!\left(\mS^k_t\vq_t\right)$ \\
\bottomrule
\end{tabular} }
\label{tab:models}
\end{table*}

\section{\ours: Persistent Memory with Sparse Routing}
\label{sec:ours}

\Cref{sec:RSM} exposes a gap in the design space: SSMs write densely and rely on decay to forget; SWA writes sparsely but is input-blind; GSA and ABC~\citep{abc} add input-dependent routing, but the writes remain dense, since every slot receives a nonzero update at each step. \ours fills this gap. It maintains separate key/value states (as in SWA), uses a \emph{sparse}, \emph{input-dependent} router to control where writes land, and retains explicit decay to control how long they persist.
Thus, relative to SWA, \ours improves memory allocation through content-dependent routing and position handling through decay rather than RoPE \citep{legacy}; relative to dense SSMs, it addresses the persistence limitations of dense state updates by writing selectively to a subset of slots.

\subsection{\ours Memory Update}

\paragraph{Recurrence.}
The central design principle of \ours is to decouple \textit{where} information is stored from \textit{how long} it persists. We implement this with a routed slot update that lets different slots be updated at different times:
\begin{equation}
\label{eq:phoenix_recurrence}
\begin{aligned}
\mS^{k}_t &= \exp(a_t \vr_t) \odot \mS^{k}_{t-1} + \bigl(\mathbf{1} - \exp(a_t \vr_t)\bigr)\vk_t^\top, \\
\mS^{v}_t &= \exp(a_t \vr_t) \odot \mS^{v}_{t-1} + \bigl(\mathbf{1} - \exp(a_t \vr_t)\bigr)\vv_t^\top,
\end{aligned}
\end{equation}
with readout $\vo_t = (\mS^{v}_t)^\top\,\softmax\bigl(\mS^{k}_t\vq_t\bigr)$, where $\odot$ broadcasts over the feature dimension $d$. When $\vr_t[i]$ is large, slot $i$ is \textit{decayed and overwritten}. When $\vr_t[i] \approx 0$, $\exp(a_t\vr_t[i]) \approx 1$ and the slot is \textit{left untouched}. \ours uses separate SWA-like states for writing keys and values.

\paragraph{Decay.}
The scalar $a_t < 0$ controls the rate of forgetting for the slots that are written to. \ours adopts Mamba--2's input-dependent per-head scalar decay in logarithmic scale~\citep{mamba2}:
\begin{equation}
\label{eq:phoenix_decay}
a_t = -\mathrm{SoftPlus}(\vw^\top\vx_t)\,\exp(\Delta).
\end{equation}

Here, $\Delta$ is a learnable scalar, as in Mamba-2. Since $a_t$ enters the recurrence only through $\exp(a_t \vr_t)$, unselected slots ($\vr_t[i] = 0$) experience no decay regardless of $a_t$; their content is frozen until the router selects them.

\paragraph{Router.}
The router determines which slots receive writes at each step. It needs to be (i) sparse, so that unselected slots are fully preserved, and (ii) input-dependent, so that the model can learn to route by content rather than by position. \ours achieves both by adapting the DeepSeek MoE routing strategy~\citep{deepseek3,deepseekmoe}: raw scores $\vm_t = \sigma(\mW\vx_t) \in \R^M$ are sparsified by keeping only the top-$K$ entries, then normalized:
\begin{equation}
\label{eq:phoenix_rout}
\vg_t = \mathrm{KeepTop}_K(\vm_t)=
\begin{cases}
\vm_t[i], & \text{if } i \in \mathrm{TopK}(\vm_t),\\
0, & \text{otherwise}
\end{cases},
\qquad
\vr_t = \frac{\vg_t}{\alpha\sum_{i=1}^M \vg_t[i]}.
\end{equation}
Here, $\alpha\in\R$ is a hyperparameter that normalizes the router as the training sequence length grows (similar to $\tau$ for GLA). We choose $\alpha=1$ for the 400M model and $\alpha=4$ for the 800M model.
This yields routing weights that sum to one over at most $K$ selected slots. Routing design choices are ablated in \Cref{subsec:router_abl}.

Unlike MoE routers, \ours omits a load-balancing loss~\citep{shazeer2017outrageously}. While uniform expert usage is desirable in MoEs, uniform memory allocation is counterproductive here: \ours intentionally routes unevenly. Retrieval-critical tokens (e.g., passkeys) are routed to dedicated slots and protected from overwrite, while general content is distributed across shared slots. The non-uniform distribution of routing results in different slots observing different sequence lengths as shown in \Cref{fig:esl1} (detailed in \Cref{sec:memory_allocation_analysis}). This mirrors the finding that recall in Transformers concentrates in a small subset of heads~\citep{bick2504understanding}; \Cref{fig:phoenix_memories} shows \ours exhibiting the same specialization at the slot level. 

\subsection{\ours Block Design}

\begin{figure}[t]
\centering
\begin{minipage}[t]{0.48\linewidth}
    \centering
    \includegraphics[width=\linewidth]{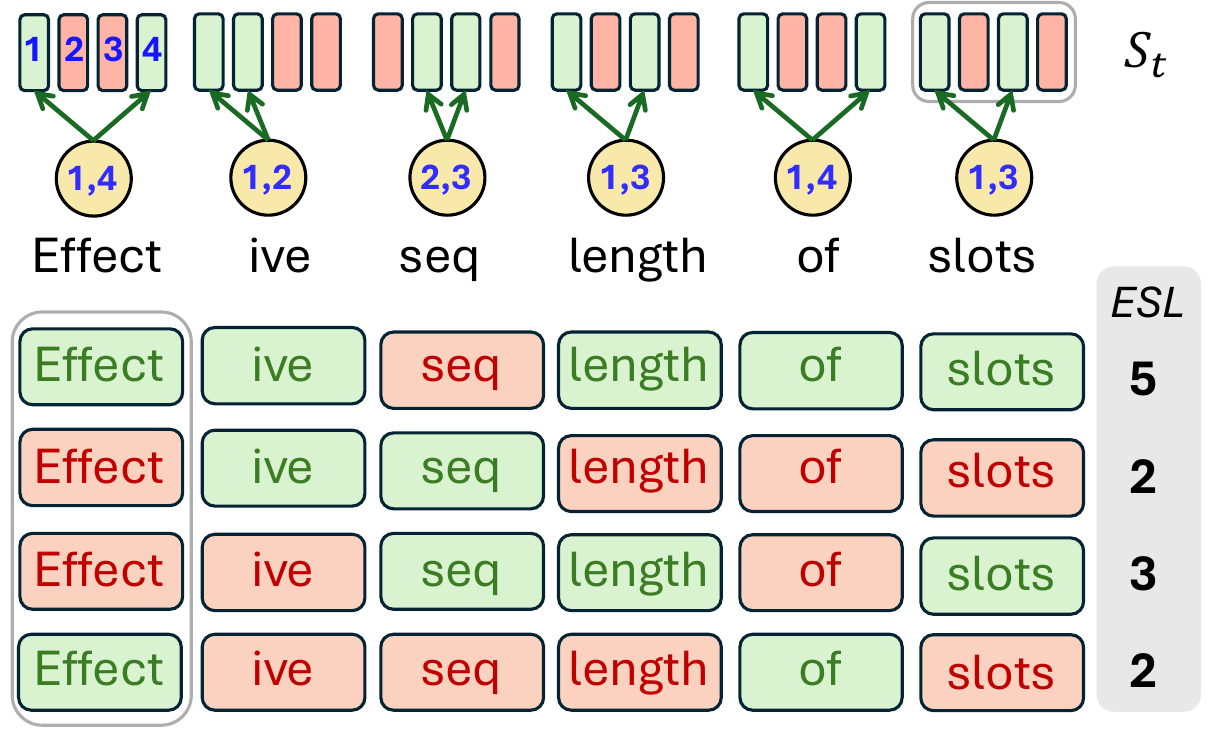}
    \captionof{figure}{\textbf{Effective Sequence Length Visualization.} For a hidden state with ${M=4}$ memory slots and a sequence of length $T=6$, each memory slot processes a different \textit{effective sequence length (\textbf{ESL})}, depending on the router.}
    \label{fig:esl1}
\end{minipage}
\hfill
\begin{minipage}[t]{0.48\linewidth}
    \centering
    \includegraphics[width=\linewidth]{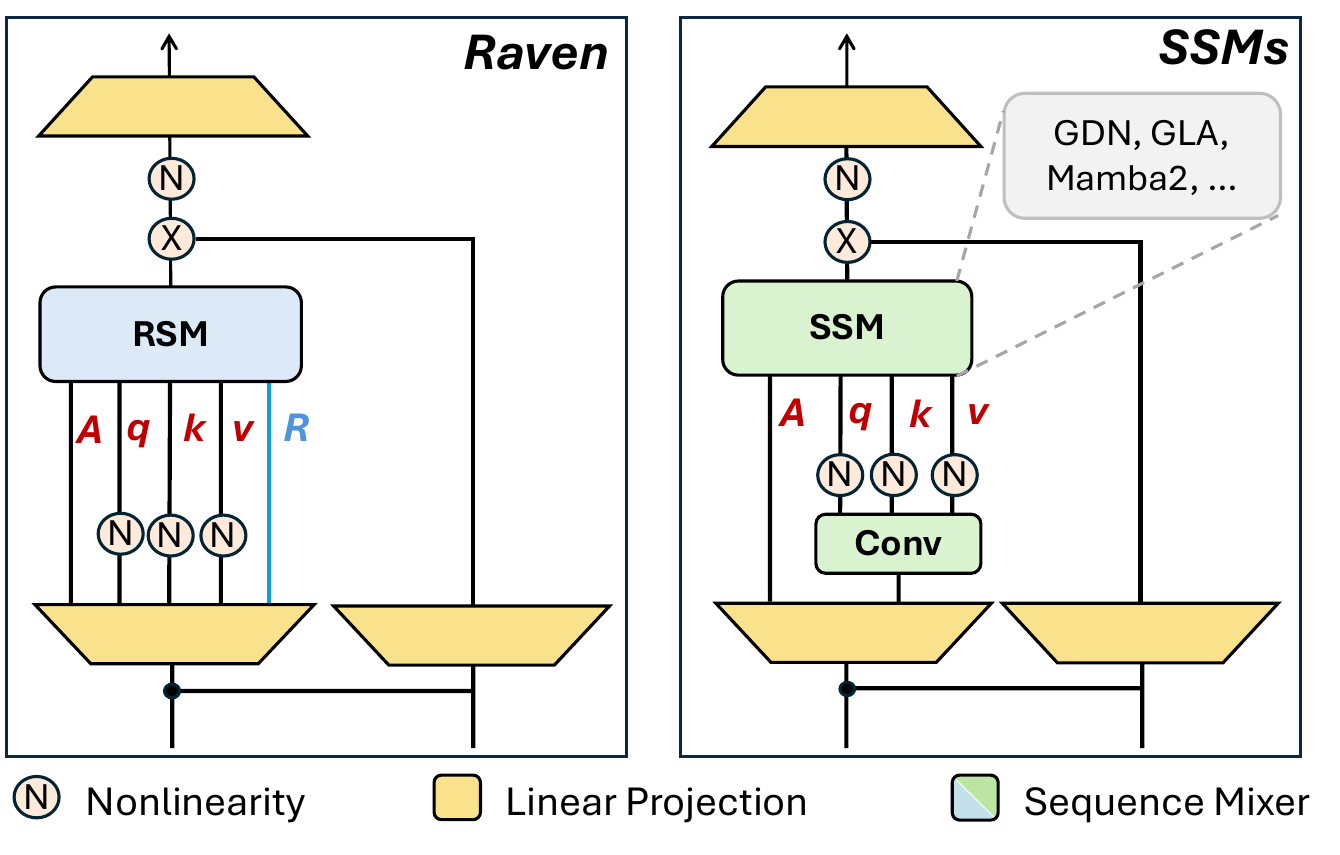}
    \captionof{figure}{\textbf{\ours Neural Architecture.} \textbf{\textit{(Left)}} \ours block vs.\ \textbf{\textit{(Right)}} other linear models. \ours requires no short-range convolutions, yielding a simpler block than existing SSMs and linear transformers.}
    \label{fig:arch_diff}
\end{minipage}
\end{figure}

\ours uses a simple block consisting of a channel mixer, standard nonlinearities, and normalization. Notably, it omits short-range convolutions entirely.

\begin{itemize}[leftmargin=*]
    \item \textbf{Channel Mixer.} \ours uses a Gated MLP~\citep{gmlp}, widely adopted in both linear~\citep{yang2024gated,deltanet} and softmax~\citep{qwen3} transformers at medium scale ($<\!10$B).
    \item \textbf{Nonlinearities and Normalization.} \ours uses SiLU activations (as in Mamba-1 and Gated DeltaNet), followed by optional QK-RMSNorm \citep{henry2020query}. 
    \item \textbf{No Short Convolutions.} Many SSMs and linear transformers apply short-range convolutions to queries, keys, and values~\citep{mamba,yang2024gated,mesanet}. \ours drops these entirely, simplifying the block without sacrificing performance. A comparison between the \ours block and other linear models is shown in \Cref{fig:arch_diff}.
\end{itemize}

\begin{table*}[t]
\centering
\caption{
\textbf{In-context recall benchmarks and NIAH accuracy vs.\ context length and cache size.}
We report accuracy (\%) on SWDE/FDA/SQuAD and on single NIAH-1/2/3 across context lengths.
Rec.\ mem.\ and Conv.\ mem.\ denote the millions of cached state elements used during decoding.
The common baseline uses $12.5$M recurrent elements in total. For the 24-layer models, this corresponds to approximately $0.52$M recurrent elements per layer; Mamba-2 distributes the same total across 48 layers.
}
\label{tab:recall_niah_merged}
\setlength{\tabcolsep}{4.5pt}
\resizebox{\textwidth}{!}{\begin{tabular}{l c c |>{\columncolor{gray!10}}c>{\columncolor{gray!10}}c>{\columncolor{gray!10}}c|cccccc|cccccc|cccccc}
\toprule
\textbf{Model} & \textbf{\# Params} & \textbf{(Rec / Conv)} &
\multicolumn{3}{c|}{\cellcolor{gray!10}\textbf{Recall tasks}} &
\multicolumn{6}{c|}{\textbf{NIAH-1}} &
\multicolumn{6}{c|}{\textbf{NIAH-2}} &
\multicolumn{6}{c}{\textbf{NIAH-3}} \\
& (M) & Mem. (M elts) &
\cellcolor{gray!10}{SWDE} &
\cellcolor{gray!10}{FDA} &
\cellcolor{gray!10}{SQuAD} &
\cellcolor{gray!10}{1K} & \cellcolor{gray!10}{2K} & \cellcolor{gray!10}{4K} & \cellcolor{gray!10}{8K} & \cellcolor{gray!10}{16K} & \cellcolor{gray!10}{32K} &
\cellcolor{gray!10}{1K} & \cellcolor{gray!10}{2K} & \cellcolor{gray!10}{4K} & \cellcolor{gray!10}{8K} & \cellcolor{gray!10}{16K} & \cellcolor{gray!10}{32K} &
\cellcolor{gray!10}{1K} & \cellcolor{gray!10}{2K} & \cellcolor{gray!10}{4K} & \cellcolor{gray!10}{8K} & \cellcolor{gray!10}{16K} & \cellcolor{gray!10}{32K} \\
\midrule
\textbf{\textit{$\sim$400M / 15B tokens}}  &  &  &
\multicolumn{3}{c|}{\cellcolor{gray!10}} &
\multicolumn{6}{c|}{} &
\multicolumn{6}{c|}{} &
\multicolumn{6}{c}{} \\
\textbf{\textit{Transformer}} &  &  &
\multicolumn{3}{c|}{\cellcolor{gray!10}} &
\multicolumn{6}{c|}{} &
\multicolumn{6}{c|}{} &
\multicolumn{6}{c}{} \\
\textit{w.\ RoPE} & 340 & $\bm \infty$ / 0.0 &
\cellcolor{gray!10}{\underline{42.3}} & \cellcolor{gray!10}{\underline{34.5}} & \cellcolor{gray!10}{\underline{22.1}} &
\textbf{100} & \textbf{100} & 0.0 & 0.0 & 0.0 & 0.0 &
\textbf{100} & \textbf{100} & 0.0 & 0.0 & 0.0 & 0.0 &
\underline{71.6} & \underline{47.6} & 0.0 & 0.0 & 0.0 & 0.0 \\
\textit{w.\ Gate (FoX)} & 376 & $\bm \infty$ / 0.0 &
\cellcolor{gray!10}\textbf{52.5} & \cellcolor{gray!10}\textbf{64.3} & \cellcolor{gray!10}\textbf{30.1} &
\textbf{100} & \textbf{100} & \textbf{32.2} & \textbf{8.0} & \textbf{4.2} & 0.0 &
\textbf{100} & \textbf{100} & \textbf{100} & \textbf{24.0} & \textbf{11.6} & \textbf{3.2} &
\textbf{95.4} & \textbf{85.6} & \textbf{64.2} & \textbf{11.6} & \textbf{7.2} & 0.0 \\
\textbf{\textit{SSM}} &  &  &
\multicolumn{3}{c|}{\cellcolor{gray!10}} &
\multicolumn{6}{c|}{} &
\multicolumn{6}{c|}{} &
\multicolumn{6}{c}{} \\
GLA & 475 & 12.5 / 0.4 &
{29.0} & {11.4} & {30.3} &
74.6 & 25.1 & 8.2 & 2.2 & 0.0 & 0.0 &
91.2 & 37.2 & 21.4 & 3.6 & 0.0 & 0.0 &
\underline{84.2} & \underline{57.1} & \underline{20.8} & \textbf{10.2} & \underline{2.3} & 0.0 \\
GSA & 399 & 12.5 / \textbf{0.0} &
\cellcolor{gray!10}{23.8} & \cellcolor{gray!10}{14.5} & \cellcolor{gray!10}{24.9} &
\underline{99.2} & \underline{97.1} & \underline{90.0} & 67.4 & 29.6 & {11.0} &
96.6 & \textbf{98.8} & 28.0 & 5.1 & 1.0 & 0.0 &
60.0 & 30.1 & 13.5 & 1.0 & 0.0 & 0.0 \\
GDN & 475 & 12.5 / 0.4 &
\cellcolor{gray!10}\underline{29.5} & \cellcolor{gray!10}{8.3} & \cellcolor{gray!10}{31.3} &
\underline{99.2} & \textbf{100} & \textbf{99.8} & \underline{92.0} & \underline{41.8} & \underline{22.1} &
\underline{99.2} & 92.0 & 43.6 & \underline{17.8} & \underline{6.2} & \underline{4.0} &
\textbf{92.6} & \textbf{80.6} & \textbf{37.8} & \underline{5.2} & \textbf{6.8} & \underline{2.5} \\
Mamba-2 & 382 & 12.5 / 0.4 &
\cellcolor{gray!10}{25.7} & \cellcolor{gray!10}\underline{14.9} & \cellcolor{gray!10}\underline{31.9} &
\underline{99.2} & 95.6 & 52.2 & 12.8 & 5.4 & 2.8 &
\textbf{99.8} & \underline{98.0} & \underline{68.2} & 15.4 & 4.4 & {3.8} &
53.4 & 53.6 & 17.4 & 1.8 & 2.2 & \textbf{3.2} \\
SWA & 374 & 12.5 / \textbf{0.0} &
\cellcolor{gray!10}{10.0} & \cellcolor{gray!10}{14.4} & \cellcolor{gray!10}{29.7} &
29.8 & 11.0 & 6.2 & 3.4 & 1.2 & 0.0 &
36.2 & 14.4 & 10.2 & 3.8 & 3.2 & 0.0 &
26.2 & 9.2 & 7.4 & 1.4 & 1.8 & 0.0 \\
\rowcolor{rmmblue}
\ours & 424 & 12.5 / \textbf{0.0} &
\textbf{34.1} &
\textbf{22.7} &
\textbf{35.4} &
\textbf{99.8} & \textbf{100} & \textbf{99.8} & \textbf{99.8} & \textbf{99.4} & \textbf{91.4} &
98.8 & \underline{98.0} & \textbf{98.8} & \textbf{81.6} & \textbf{23.0} & \textbf{8.8} &
76.8 & 43.6 & 13.4 & 1.0 & 0.0 & 0.0 \\
\midrule
\textbf{\textit{$\sim$800M / 32B tokens}}  &  &  &
\multicolumn{3}{c|}{\cellcolor{gray!10}} &
\multicolumn{6}{c|}{} &
\multicolumn{6}{c|}{} &
\multicolumn{6}{c}{} \\
\textbf{\textit{Transformer}} &  &  &
\multicolumn{3}{c|}{\cellcolor{gray!10}} &
\cellcolor{gray!10}{2K} & \cellcolor{gray!10}{4K} & \cellcolor{gray!10}{8K} & \cellcolor{gray!10}{16K} & \cellcolor{gray!10}{32K} & \cellcolor{gray!10}{64K} &
\cellcolor{gray!10}{2K} & \cellcolor{gray!10}{4K} & \cellcolor{gray!10}{8K} & \cellcolor{gray!10}{16K} & \cellcolor{gray!10}{32K} & \cellcolor{gray!10}{64K} &
\cellcolor{gray!10}{2K} & \cellcolor{gray!10}{4K} & \cellcolor{gray!10}{8K} & \cellcolor{gray!10}{16K} & \cellcolor{gray!10}{32K} & \cellcolor{gray!10}{64K} \\
\textit{w. RoPE} & 693 & $\bm \infty$ / 0.0 &
\cellcolor{gray!10}{\underline{58.9}} & \cellcolor{gray!10}{\underline{63.6}} & \cellcolor{gray!10}{\underline{41.3}} &
\underline{92.4} & \underline{90.8} & 0.0 & 0.0 & 0.0 & 0.0 &
\underline{99.6} & \underline{99.2} & 0.0 & 0.0 & 0.0 & 0.0 &
\underline{33.4} & \underline{9.0} & 0.0 & 0.0 & 0.0 & 0.0 \\
\textit{w. Gate (FoX)} & 694 & $\bm \infty$ / 0.0 &
\cellcolor{gray!10}{\textbf{64.9}} & \cellcolor{gray!10}{\textbf{80.0}} & \cellcolor{gray!10}{\textbf{41.5}} &
\textbf{100} & \textbf{100} & \textbf{99.0} & \textbf{28.6} & \textbf{12.4} & 0.0 &
\textbf{100} & \textbf{100} & \textbf{67.4} & \textbf{9.0} & \textbf{7.8} & 0.0 &
\textbf{83.2} & \textbf{63.8} & \textbf{0.2} & \textbf{0.4} & \textbf{0.2} & 0.0 \\
\textbf{\textit{SSM}} &  &  &
\multicolumn{3}{c|}{\cellcolor{gray!10}} &
\multicolumn{6}{c|}{} &
\multicolumn{6}{c|}{} &
\multicolumn{6}{c}{} \\
GLA & 892 & 16.5 / 0.4 &
\cellcolor{gray!10}{\textbf{50.1}} & \cellcolor{gray!10}{\textbf{35.9}} & \cellcolor{gray!10}{\underline{39.2}} &
100 & 94.2 & 29.4 & 3.6 & 0.0 & 0.0 &
100 & 89.8 & 25.8 & 3.8 & 3.0 & 0.0 &
94.6 & 51.2 & 3.2 & 1.8 & 1.6 & 0.0 \\
GSA & 750 & 16.5 / \textbf{0.0} &
\cellcolor{gray!10}{42.8} & \cellcolor{gray!10}{28.9} & \cellcolor{gray!10}{33.6} &
100 & 100 & 99.8 & 97.6 & 64.0 & 0.0 &
99.4 & 90.0 & 29.8 & 3.6 & 2.4 & 0.0 &
79.0 & 16.6 & 2.4 & 0.2 & 0.0 & 0.0 \\
GDN & 892 & 16.5 / 0.4 &
\cellcolor{gray!10}{\underline{46.0}} & \cellcolor{gray!10}{\underline{31.9}} & \cellcolor{gray!10}{35.9} &
\textbf{100} & \textbf{100} & \textbf{100} & \textbf{100} & 94.8 & 45.2 &
\textbf{100} & 95.4 & 35.2 & 14.2 & \textbf{9.2} & \textbf{5.8} &
79.4 & \textbf{65.4} & 10.6 & 6.4 & \textbf{4.2} & \textbf{2.0} \\
Mamba-2 & 711 & 16.5 / 0.6 &
\cellcolor{gray!10}{42.7} & \cellcolor{gray!10}{21.3} & \cellcolor{gray!10}{34.6} &
99.8 & 63.4 & 13.6 & 1.8 & 0.8 & 0.6 &
97.4 & 49.2 & 24.2 & 6.6 & 4.2 & 3.2 &
89.8 & 17.4 & \textbf{14.6} & 1.4 & 2.4 & 0.6 \\
SWA & 693 & 16.5 / \textbf{0.0} &
\cellcolor{gray!10}{14.9} & \cellcolor{gray!10}{12.2} & \cellcolor{gray!10}{28.8} &
11.0 & 6.2 & 3.4 & 1.2 & 0.6 & 0.0 &
14.4 & 6.4 & 5.0 & 1.2 & 2.4 & 0.0 &
15.2 & 4.8 & 1.4 & 2.4 & 3.4 & 0.0 \\
\rowcolor{rmmblue}
\ours & 792 & 16.5 / \textbf{0.0} &
41.7 & 24.1 & \textbf{39.9} &
99.8 & \textbf{100} & 99.8 & 99.2 & \textbf{98.2} & \textbf{91.0} &
\textbf{100} & \textbf{99.4} & \textbf{90.8} & \textbf{35.2} & 6.8 & 3.4 &
\textbf{90.4} & 5.2 & 0.6 & 0.0 & 0.0 & 0.0 \\
\bottomrule
\end{tabular}}
\end{table*}

\subsection{Unifying Sparse and Selective Writes}
\label{gdnunif}

The recurrence in \Cref{eq:phoenix_recurrence} is specific to \ours, but the underlying write-and-forget structure it embodies is shared across a broader family of models. Despite their differences in decay, readout, and state structure, SWA and DeltaNet can both be written as:
\begin{equation*}
\mS_t \;=\; \mS_{t-1}(\mI - \vr_t \vr_t^\top) + \vv_t \vr_t^\top.
\end{equation*}
SWA instantiates this with a one-hot write vector $\vr_t = \ve_t$ (sparse, full overwrite of a single slot); DeltaNet uses $\vr_t = \vk_t$ (dense, distributed update across all slots).
\ours occupies the middle ground between both, using a diagonal overwrite matrix $\mP_t$ rather than a rank-one projector $\vr_t\vr_t^\top$:
\begin{equation}
\mS_t \;=\; \mS_{t-1}\,(\mI - \mP_t) + \vv_t\,\vp_t^\top,
\label{eq:routing_spectrum}
\end{equation}
with the following instantiations:
\[
\begin{array}{rll}
\textbf{SWA:}      & \vp_t = \ve_t,                        & \mP_t = \ve_t\ve_t^\top, \\[0.3em]
\textbf{\ours:}    & \vp_t = \mathbf{1}-\exp(a_t\vr_t),   & \mP_t = \mathrm{diag}(\vp_t), \\[0.3em]
\textbf{DeltaNet:} & \vp_t = \vk_t,                        & \mP_t = \vk_t\vk_t^\top.
\end{array}
\]
\ours occupies the middle ground: $\mP_t$ is diagonal (not rank-one), sparse (at most $K$ nonzero entries), and input-dependent. This enables slot-wise, selective overwrites---recovering SWA's sparsity and SSMs' decay within a single unified update.

\section{Empirical Validation}
\label{empirical}

\begin{table}[t]
\centering
\caption{
\textbf{Zero-shot language modeling performance across models.}
Left: 400M models trained on 15B tokens. 
Right: 800M models trained on 32B tokens.
}

\small
\setlength{\tabcolsep}{2.0pt}
\renewcommand{\arraystretch}{1.05}

\begin{minipage}[t]{0.48\columnwidth}
\centering
\textbf{\textit{400M / 15B tokens}}

\vspace{3pt}

\resizebox{\linewidth}{!}{\begin{tabular}{lc|ccccccc|>{\columncolor{orange!10}}c}
\toprule
\textbf{Model} & \textbf{\# Params} & \textbf{LMB.} & \textbf{LMB.} & \textbf{PIQA} & \textbf{Hella.} & \textbf{Wino.} & \textbf{ARC-e} & \textbf{ARC-c} & \textbf{Avg.} \\
& (M) & ppl$\downarrow$ & acc$\uparrow$ & acc$\uparrow$ & acc$\uparrow$ & acc$\uparrow$ & acc$\uparrow$ & acc$\uparrow$ & acc$\uparrow$ \\
\midrule

\multicolumn{10}{l}{\textbf{\textit{Transformer}}} \\
\textit{w.\ RoPE} & 340 & 42.0 & 31.0 & 64.4 & 30.2 & 51.0 & 44.3 & 18.7 & 39.9 \\
\textit{w.\ Gate (FoX)} & 376 & 48.1 & 30.6 & 64.9 & 30.7 & 51.1 & 44.7 & 18.9 & 40.1 \\

\multicolumn{10}{l}{\textbf{\textit{SSM}}} \\
GLA   & 475 & 42.1 & 30.7 & 64.4 & 30.1 & 52.7 & 43.8 & 19.6 & 40.2 \\
GSA   & 399 & 44.1 & 30.3 & 64.9 & 30.7 & 51.5 & 45.6 & 20.5 & \underline{40.5} \\
GDN   & 475 & {40.1} & 31.6 & 65.6 & 31.4 & 50.2 & 45.7 & 19.3 & \textbf{40.6} \\
Mamba-2 & 382 & 43.0 & 29.9 & 65.0 & 31.5 & 51.2 & 47.5 & 20.5 & 40.1 \\
SWA   & 374 & 40.7 & 30.5 & 64.5 & 30.4 & 51.6 & 44.9 & 18.6 & 40.0 \\

\rowcolor{rmmblue}
\ours  & 424 & 41.0 & 32.7 & 64.1 & 30.3 & 51.7 & 43.9 & 18.4 & 40.2 \\

\bottomrule
\end{tabular}}
\end{minipage}
\hfill
\begin{minipage}[t]{0.48\columnwidth}
\centering
\textbf{\textit{800M / 32B tokens}}

\vspace{3pt}

\resizebox{\linewidth}{!}{\begin{tabular}{lc|ccccccc|>{\columncolor{orange!10}}c}
\toprule
\textbf{Model} & \textbf{\# Params} & \textbf{LMB.} & \textbf{LMB.} & \textbf{PIQA} & \textbf{Hella.} & \textbf{Wino.} & \textbf{ARC-e} & \textbf{ARC-c} & \textbf{Avg.} \\
& (M) & ppl$\downarrow$ & acc$\uparrow$ & acc$\uparrow$ & acc$\uparrow$ & acc$\uparrow$ & acc$\uparrow$ & acc$\uparrow$ & acc$\uparrow$ \\
\midrule

\multicolumn{10}{l}{\textbf{\textit{Transformer}}} \\
\textit{w.\ RoPE} & 693 & {18.6} & 41.4 & 66.3 & 34.3 & 52.2 & 49.9 & 21.9 & \underline{44.5} \\
\textit{w.\ Gate (FoX)} & 694 & 25.3 & 38.2 & 67.8 & 34.4 & 51.7 & 49.9 & 21.5 & 44.1 \\

\multicolumn{10}{l}{\textbf{\textit{SSM}}} \\
GLA & 892 & 23.6 & 38.2 & 66.9 & 33.4 & 52.4 & 48.5 & 21.2 & 43.5 \\
GSA & 750 & 27.4 & 34.7 & 66.3 & 32.1 & 51.6 & 46.6 & 19.6 & 41.8 \\
GDN & 892 & 21.3 & 39.4 & 68.1 & 35.2 & 53.0 & 52.5 & 22.1 & \textbf{45.1} \\
Mamba-2 & 712 & 24.6 & 36.0 & 68.1 & 35.4 & 52.6 & 52.3 & 22.3 & \underline{44.5} \\
SWA & 693 & 20.8 & 38.8 & 67.9 & 34.2 & 52.8 & 49.3 & 21.2 & 44.0 \\

\rowcolor{rmmblue}
\ours & 792 & 26.0 & 38.2 & 67.0 & 33.2 & 50.9 & 49.2 & 21.0 & 43.3 \\

\bottomrule
\end{tabular}}
\end{minipage}

\vspace{3pt}
\label{tab:language_modeling}
\end{table}

We evaluate \ours across three axes: retrieval ability, general language modeling, and hybrid performance. Together, these experiments validate that sparse, input-dependent routing improves memory utilization in ways that translate across tasks, context lengths, and architectural configurations.

\paragraph{Experimental Setup.}
We follow the standard training recipe used across linear and quadratic sequence models, carefully matching memory size and parameter counts across baselines to ensure fair comparison. Full training details and model configurations are provided in \Cref{app:experimental_details}.

\begin{figure}[t]
    \centering
    \includegraphics[width=1.0\linewidth]{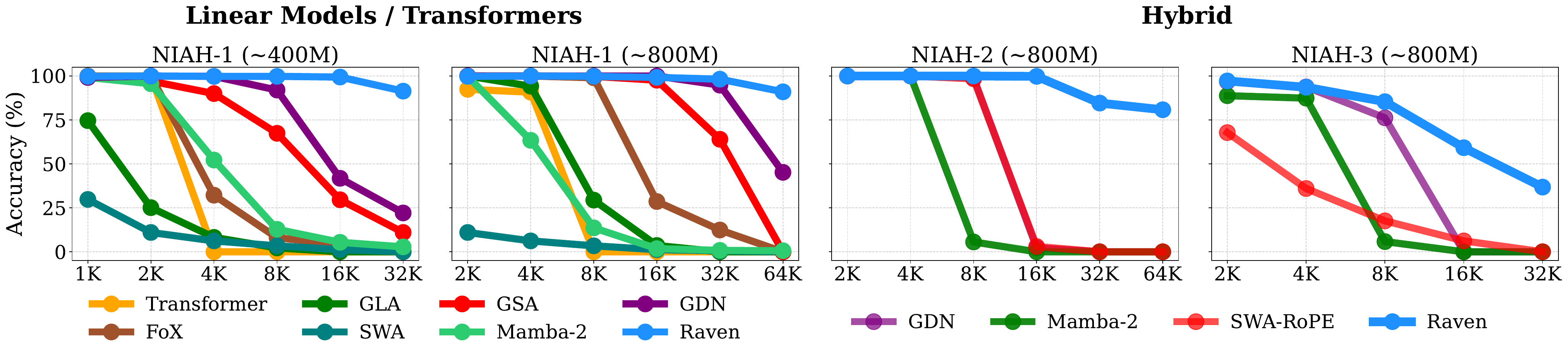}
\caption{
\textbf{NIAH Results.}
\textbf{\textit{(Left)}} NIAH-1 for 400M and 800M SSMs and transformers.
\textbf{\textit{(Right)}} NIAH-2 and NIAH-3 for 800M hybrids.
}
\label{fig:niah_results}
\end{figure}

\subsection{Retrieval Abilities}
\label{subsec:retrieval_abilities}

\begin{table*}[t]
\centering
\caption{
\textbf{Hybrid Models Retrieval Ability.} We evaluate GDN, Mamba-2, SWA-RoPE, and
\ours{} as the linear component in a hybrid setting with full attention. Attention blocks use NoPE positional embeddings. We report performance on NIAH-1, NIAH-2, NIAH-3 and on $\hgray{\text{highlighted}}$ SWDE, FDA, and SQuAD. The 400M models are trained on 2K sequences, and the 800M models are trained on 4K sequences. Full ablations appear in \Cref{tab:full_hybrid_comparison}.
}
\resizebox{\textwidth}{!}{
\begin{tabular}{lc | >{\columncolor{gray!10}}c>{\columncolor{gray!10}}c>{\columncolor{gray!10}}c |
cccccc | cccccc | cccccc}
\toprule
\textbf{Linear} & \textbf{No} &
\textbf{SWDE} & \textbf{FDA} & \textbf{SQuAD} &
\multicolumn{6}{c|}{\textbf{NIAH-1}} &
\multicolumn{6}{c|}{\textbf{NIAH-2}} &
\multicolumn{6}{c}{\textbf{NIAH-3}} \\
\textbf{Model} & \textbf{Conv.} &
acc$\uparrow$ & acc$\uparrow$ & acc$\uparrow$ &
\cellcolor{gray!10}{1K} & \cellcolor{gray!10}{2K} & \cellcolor{gray!10}{4K} & \cellcolor{gray!10}{8K} & \cellcolor{gray!10}{16K} & \cellcolor{gray!10}{32K} &
\cellcolor{gray!10}{1K} & \cellcolor{gray!10}{2K} & \cellcolor{gray!10}{4K} & \cellcolor{gray!10}{8K} & \cellcolor{gray!10}{16K} & \cellcolor{gray!10}{32K} &
\cellcolor{gray!10}{1K} & \cellcolor{gray!10}{2K} & \cellcolor{gray!10}{4K} & \cellcolor{gray!10}{8K} & \cellcolor{gray!10}{16K} & \cellcolor{gray!10}{32K} \\
\midrule
{\textbf{\textit{$\sim$ 400M / 15B tokens}}} & & \cellcolor{gray!10}{} & \cellcolor{gray!10}{} & \cellcolor{gray!10}{} &
& & & & & &
& & & & & &
& & & & & \\
GDN & \textcolor{red}{\xmark}
& \underline{54.6} & 67.2 & \underline{34.5}
& \textbf{100} & \textbf{100} & \textbf{100} & \textbf{100} & 93.2 & 70.5
& \textbf{100} & \textbf{100} & \textbf{100} & 8.0 & 0.0 & 0.0
& 93.2 & 70.2 & 50.0 & 0.0 & 0.0 & 0.0 \\
Mamba-2 & \textcolor{red}{\xmark}
& \textbf{56.3} & \textbf{68.8} & \textbf{36.0}
& \textbf{100} & \textbf{100} & 16.4 & 0.0 & 0.0 & 0.0
& \textbf{100} & \textbf{100} & 85.8 & 0.0 & 0.0 & 0.0
& 76.9 & \textbf{80.6} & 60.8 & 0.0 & 0.0 & 0.0 \\
SWA-RoPE & \greencheck
& 51.0 & \underline{68.1} & 34.1
& \textbf{100} & \textbf{100} & \textbf{100} & \textbf{100} & 98.2 & 60.4
& \textbf{100} & \textbf{100} & \textbf{100} & 98.2 & 3.1 & 0.0
& \textbf{93.4} & 78.2 & 12.8 & \textbf{60.0} & 4.4 & 0.0 \\
\rowcolor{rmmblue}
\ours & \greencheck
& 51.4 & 64.2 & 31.4
& \textbf{100} & \textbf{100} & \textbf{100} & \textbf{100} & 98.4 & \textbf{78.6}
& \textbf{100} & \textbf{100} & \textbf{100} & \textbf{100} & \textbf{95.4} & \textbf{65.4}
& 90.0 & 67.0 & \textbf{73.8} & \textbf{60.0} & \textbf{10.2} & \textbf{14.4} \\
\midrule
{\textbf{\textit{$\sim$ 800M / 32B tokens}}} & & \cellcolor{gray!10}{} & \cellcolor{gray!10}{} & \cellcolor{gray!10}{} &
\cellcolor{gray!10}{2K} & \cellcolor{gray!10}{4K} & \cellcolor{gray!10}{8K} & \cellcolor{gray!10}{16K} & \cellcolor{gray!10}{32K} & \cellcolor{gray!10}{64K} &
\cellcolor{gray!10}{2K} & \cellcolor{gray!10}{4K} & \cellcolor{gray!10}{8K} & \cellcolor{gray!10}{16K} & \cellcolor{gray!10}{32K} & \cellcolor{gray!10}{64K} &
\cellcolor{gray!10}{2K} & \cellcolor{gray!10}{4K} & \cellcolor{gray!10}{8K} & \cellcolor{gray!10}{16K} & \cellcolor{gray!10}{32K} & \cellcolor{gray!10}{64K} \\
GDN & \textcolor{red}{\xmark}
& \underline{64.7} & \underline{77.8} & \textbf{48.1}
& \textbf{100} & \textbf{100} & \textbf{100} & 63.4 & 0.2 & 0.0
& \textbf{100} & \textbf{100} & \textbf{99.0} & 2.2 & 0.0 & 0.0
& 96.8 & 93.8 & 76.2 & 0.0 & 0.0 & 0.0 \\
Mamba-2 & \textcolor{red}{\xmark}
& \textbf{68.5} & 72.9 & \underline{42.3}
& \textbf{100} & \textbf{100} & 0.0 & 0.0 & 0.0 & 0.0
& \textbf{100} & \textbf{100} & 5.6 & 0.0 & 0.0 & 0.0
& 88.8 & 87.4 & 5.8 & 0.0 & 0.0 & 0.0 \\
SWA-RoPE & \greencheck
& 63.4 & 68.5 & 17.2
& \textbf{100} & \textbf{100} & \textbf{100} & \textbf{100} & 97.6 & \textbf{69.0}
& \textbf{100} & \textbf{100} & 98.4 & 3.0 & 0.0 & 0.0
& 67.8 & 36.0 & 17.6 & 6.2 & 0.0 & 0.0 \\
\rowcolor{rmmblue}
\ours & \greencheck
& {64.5} & \textbf{81.3} & 37.0
& \textbf{100} & \textbf{100} & \textbf{100} & 99.8 & \textbf{99.4} & 55.4
& \textbf{100} & \textbf{100} & \textbf{100} & \textbf{99.8} & \textbf{84.6} & \textbf{80.8}
& \textbf{97.2} & \textbf{93.6} & \textbf{85.4} & \textbf{59.2} & \textbf{36.8} & 0.0 \\
\bottomrule
\end{tabular}
}
\label{tab:hybrid_comparison_selected}
\end{table*}

\paragraph{Single Needle-in-a-Haystack (NIAH).}
We evaluated passkey retrieval across varying depths and sequence lengths~\citep{ruler}, with 400M models trained at a context length of 2048 tokens and 800M models trained at a context length of 4096. The results in \Cref{tab:recall_niah_merged} are stark. Strong baselines like Mamba-2 and GDN degrade significantly beyond 8K tokens in the 400M range --- already $4\times$ their training length, as their dense state update forces every slot to compress the full token history, rapidly diluting stored information. \ours maintains near-perfect accuracy ($\geq\mathbf{99\%}$) up to 16K tokens and is the \textit{only model} at the 400M scale to retain strong performance ($>\mathbf{91\%}$) at 32K, $\mathbf{16\times}$ its training length. This is a direct consequence of sparse routing: by writing selectively, \ours prevents memory slots from being overloaded and naturally extrapolates to longer contexts without any explicit length curriculum. As shown in \Cref{fig:niah_results}, \ours achieves the strongest performance on the NIAH-1 task, attaining near-perfect recall relative to both linear models and softmax transformers at both parameter scales. At the 800M scale, \ours demonstrates the strongest length generalization beyond its training sequence length, retaining $\mathbf{91.0\%}$ on NIAH-1 at 64K where the next-best baseline (GDN) drops to $45.2\%$. Furthermore, \ours surpasses even strong Transformer models such as FoX~\citep{fox} on NIAH-1, owing to its exceptional length generalization ability.

\paragraph{Recall-Intensive Benchmarks.}
Beyond synthetic retrieval, we evaluate on real-world recall-heavy tasks: single-document extractive QA (SQuAD), web data extraction (SWDE), and document-level information extraction (FDA). As shown in \Cref{tab:recall_niah_merged}, \ours consistently outperforms linear-time baselines across all three. On SWDE, \ours reaches \textbf{34.1\%} accuracy, surpassing Mamba-2 (25.7\%) and GLA (29.0\%) and narrowing the gap to the Transformer upper bound among 400M models. On FDA, \ours is the only linear model to exceed 22\%, improving by ${\sim}8$ points over the best SSM baseline. These gains go beyond synthetic passkey retrieval: selective forgetting preserves information needed for extraction and document understanding, not just token-level recall. At the 800M scale, \ours remains clearly competitive with the best linear Transformers, such as GDN. Softmax-based Transformers such as FoX store all tokens in memory and, unlike SSMs, do not rely on fixed-size memory. In \Cref{sec:hybrid}, we show that hybrid models match or outperform these strong Transformers, even when replacing half of the softmax layers with linear layers.

\paragraph{Retrieval Without Convolutions.}
Many recent linear models --- including GLA, Mamba-2, and GDN --- rely on 1D convolutions over token features (often applied to projected $\mQ/\mK/\mV$ streams) for local context integration and training stability. \ours attains state-of-the-art retrieval \textit{without them}, suggesting that input-dependent routing and per-slot decay are sufficient to address the recall bottleneck: the first controls \textit{where} information is written, and the second controls \textit{how long} it persists.

\subsection{Language Modeling}

The retrieval gains of \ours are only meaningful if they do not come at the cost of general language modeling quality. A model that achieves long-context retrieval by sacrificing compression ability would be of limited practical value. \Cref{tab:language_modeling} compares \ours against Mamba-2, GLA, GDN, and strong Transformer baselines including FoX~\citep{fox} at both 400M and 800M parameter scales across standard zero-shot benchmarks.

At 400M parameters, \ours matches or exceeds Mamba-2 and GLA on average accuracy while achieving the best Lambada accuracy among all models, including Transformers. At 800M, \ours trails GDN on average accuracy, though GDN uses significantly more parameters (892M vs.\ 792M). Across both scales, the introduction of sparse routing and input-dependent forgetting does not measurably degrade general language modeling; the two objectives are compatible, not in tension.

\ours provides a substantial improvement over SWA and related variants such as GSA, as shown in \Cref{tab:recall_niah_merged}. It is the only model that both selectively forgets historical information—thereby encoding the relative positions of tokens—and selectively routes each token to specific parts of its memory through input-dependent routing.

\subsection{Hybrid \ours} \label{sec:hybrid}

Hybrid architectures that interleave linear layers with attention have become a standard design pattern \citep{blakeman2025nemotron}, combining the inference efficiency of linear recurrences with the expressiveness of exact attention at critical layers. We evaluate \ours as a drop-in linear component paired with two full NoPE attention layers. We further ablate alternative configurations using RoPE as positional embeddings and SWA instead of full attention, resulting in four alternatives, which are evaluated in \Cref{tab:full_hybrid_comparison}. In this section, we primarily focus on hybrid models using full NoPE attention, as this is the most informative configuration.

\paragraph{Hybrid \ours vs. SSMs} When paired with full NoPE attention, \ours dramatically outperforms GDN and Mamba-2 hybrids on long-context NIAH. GDN+Attn degrades sharply on NIAH-2 and NIAH-3 beyond 4K, and Mamba-2+Attn collapses almost entirely past 2K on NIAH-1, whereas \ours+Attn maintains strong accuracy up to 32K on NIAH-1 and 16K on NIAH-2. 

The two components play complementary roles: NoPE attention enables precise short-range retrieval, whereas \ours’s persistent slots retain long-range information across the sequence. As shown in \Cref{tab:full_hybrid_comparison}, hybrid models match or outperform strong Transformer baselines such as FoX (presented in \Cref{tab:recall_niah_merged}) on recall-heavy tasks, particularly when evaluating beyond their training context length. Among all models, the \ours hybrid with NoPE attention performs best in this regime. \Cref{fig:niah_results} clearly demonstrates the strength of \ours. While all other hybrid approaches fail to achieve meaningful accuracy on NIAH-2 for long sequences (32K and 16K), \ours attains over $80\%$ accuracy on these tasks.

\paragraph{Hybrid \ours vs. SWA-RoPE}
Another popular hybrid configuration combines SWA with RoPE positional embeddings with full softmax NoPE attention. In this setting, SWA-RoPE is used as the linear component for efficiency instead of SSMs. As discussed in \Cref{sec:ours}, \ours improves on SWA in two ways: it replaces the fixed ring buffer with selective, input-dependent routing, and it replaces RoPE-based position handling with decay that implicitly encodes relative position.

These changes lead to large gains when \ours is used as the linear component in hybrid models. As shown in \Cref{tab:hybrid_comparison_selected}, the \ours hybrid consistently outperforms SWA-RoPE on both recall-intensive benchmarks and NIAH. In particular, it substantially improves length generalization, alleviating the main limitation of SWA-based hybrids.

\section{Ablations}
\label{sec:ablations}

We ablate the key design choices in \ours across four axes: architectural components, router design, memory shape, and routing sparsity. Together, these experiments validate the core design decisions and provide guidance on how to configure \ours for different settings.

\begin{table*}[t]
\centering

\begin{minipage}[t]{0.49\textwidth}
\centering
\vspace{0pt}
\small
\captionof{table}{\textbf{Architectural Components Effect.} Ablation results of different architectural choices of \ours. Results are for 256 memory slots with Top$_{32}$ selected by sigmoid router. Models are trained with 400M parameters on 15B tokens. \textbf{Lmb.} is lambada perplexity on evaluation set and \textbf{Avg.} shows the average performance on language modeling tasks also shown in \Cref{tab:language_modeling}.
}
\resizebox{\linewidth}{!}{\begin{tabular}{l|c>{\columncolor{orange!10}}c| >{\columncolor{gray!10}}c>{\columncolor{gray!10}}c>{\columncolor{gray!10}}c}
\toprule
\textbf{Model Ablation}& \textbf{Lmb.} &\textbf{Avg.} & \textbf{SWDE} & \textbf{FDA} & \textbf{SQuAD} \\
 & ppl  $\downarrow$& acc $\uparrow$ & acc $\uparrow$ & acc $\uparrow$ & acc $\uparrow$ \\
\midrule
\rowcolor{rmmblue}
\ours & \underline{43.1} & \textbf{40.1} & \underline{31.5} & \textbf{19.6} & \textbf{36.6} \\
\textit{w/o}. Output Gate & 48.2 & 39.1 & 29.6 & 14.6 & 30.2 \\
\textit{w}. $\mA_t = \sigma(\mW\vx_t)^{1/4}$ & 43.2 & 39.5 & 28.0 & 11.5 & 29.4 \\
\textit{w}. \textit{Short Conv} & \textbf{42.7} & \underline{40.1} & \textbf{32.2} & \underline{16.1} & 36.4 \\
\bottomrule
\end{tabular}}
\label{tab:arch_abl_component}
\end{minipage}
\hfill
\begin{minipage}[t]{0.49\textwidth}
\centering
\vspace{0pt}
\small
\captionof{table}{\textbf{Router Design Ablations.}
Router design effect on recall-heavy tasks. 
All models have 256 slots with Top$_{32}$. $\hblue{\text{{Blue}}}$ highlights the default setup used for \ours. Results for 400M parameter models.}
\resizebox{\linewidth}{!}{\begin{tabular}{l c l | >{\columncolor{gray!10}}c >{\columncolor{gray!10}}c >{\columncolor{gray!10}}c }
\toprule
\multirow{2}{*}{\makecell{\textbf{Router}\\\textbf{Type}}} &
\multirow{2}{*}{\makecell{\textbf{Gumbel}\\\textbf{Noise}}} &
\multirow{2}{*}{\makecell{\textbf{Router}\\\textbf{Projection}}} &
\textbf{SWDE} & \textbf{FDA} & \textbf{SQuAD} \\
& & & acc$\uparrow$ & acc$\uparrow$ & acc$\uparrow$ \\
\midrule
\rowcolor{rmmblue}
$\text{Sigmoid}$ & \greencheck  & Linear & \textbf{34.9} & 13.8 & \textbf{34.3} \\
$\text{Sigmoid}$ & \textcolor{red}{\xmark} & Linear & 29.7 & 17.7 & 29.5 \\
$\softmax$ & \greencheck  & Linear & 25.4 & 18.2 & 27.6 \\
$\softmax$ & \textcolor{red}{\xmark} & Linear & 26.2 & 9.4 & 25.6 \\
\midrule
$\text{Sigmoid}$ & \greencheck  & MLP & \underline{30.0} & \textbf{25.7} & \underline{30.7} \\
$\text{Sigmoid}$ & \textcolor{red}{\xmark} & MLP & \underline{30.0} & 13.3 & 20.5 \\
$\softmax$ & \greencheck  & MLP & 22.7 & \underline{21.3} & 29.2 \\
$\softmax$ & \textcolor{red}{\xmark} & MLP & 20.8 & 4.5 & 24.0 \\
\bottomrule
\end{tabular}}
\label{tab:router_ablations}
\end{minipage}

\end{table*}

\subsection{Block Components}

We ablate three architectural components common in linear and softmax transformers: output gating, decay parametrization, and short-range convolutions. Results are summarized in \Cref{tab:arch_abl_component}.

Output gating consistently improves retrieval performance, consistent with its role in prior models~\citep{deltanet,mamba2} (see \Cref{fig:arch_diff}). For decay, the temperature-scaled sigmoid parametrization used in GLA/KDA~\citep{kimilinear} underperforms \ours's decay, suggesting that Mamba-2-style input-dependent scalar decay is better suited to the routed setting. Finally, adding short-range convolutions over queries, keys, and values yields marginal gains on some benchmarks but does not change the overall picture: \ours maintains strong recall without them, unlike prior linear models~\citep{deltanet,kexuefm}. We omit convolutions in the final design for simplicity.

\subsection{Router Design}
\label{subsec:router_abl}

The router is the component of \ours with the least precedent in prior sequence modeling work; standard SSMs and linear transformers have no notion of input-dependent slot selection. The closest analog in the literature is the MoE router, which similarly maps a token to a sparse subset of modules. We therefore draw inspiration from MoE design choices and adapt them to the memory routing setting. Below we ablate the three main degrees of freedom: the scoring function, the use of stochastic exploration during training, and the expressiveness of the projection. Results are in \Cref{tab:router_ablations}, evaluated on recall-heavy benchmarks where routing quality matters most.

\subsubsection{Softmax vs.\ Sigmoid Routing}
\label{sec:r1}

The scoring function determines how slot scores are computed before Top-$K$ selection. The two standard choices in MoE architectures are softmax and sigmoid.

\textbf{Softmax} routing normalizes scores globally across all $M$ slots before selection:
\begin{equation}
\label{eq:softmaxrout}
\vm_t = \softmax(\mW\vx_t) \in \R^M.
\end{equation}
Global normalization means that the scores of unselected slots influence the selected weights, coupling routing decisions across the full slot set. This competitive structure is natural in MoEs --- where experts should not all fire at once --- but is less appropriate for memory, where independent slot assessment is preferable.

\textbf{Sigmoid} routing (\Cref{eq:phoenix_rout}) scores each slot independently~\citep{deepseekmoe}. Slots do not compete with each other, so the router can assign high confidence to multiple slots simultaneously without suppressing the rest. Importantly, sigmoid scores are naturally sharp: many slots receive values close to zero without any explicit sparsity constraint, meaning Top-$K$ selection mostly confirms what the router has already decided rather than imposing an artificial cutoff. Sigmoid routing outperforms softmax across most tasks (\Cref{tab:router_ablations}), and we adopt it as the default.

\subsubsection{Gumbel Noise During Training}
\label{sec:r2}

Without intervention, routers can collapse onto a small subset of slots early in training and never recover --- a well-known failure mode in MoEs~\citep{shazeer2017outrageously}. In our setting, router collapse is particularly harmful: slots that are never selected remain uninformative, permanently reducing the effective memory capacity of the model. To prevent this, we inject Gumbel noise~\citep{jang2016categorical} into the router logits during training:
\begin{align}
\label{eq:gumbelnoise}
\vm_t = f(\mW\vx_t + \mathbf{n}), \qquad \mathbf{n} \in \R^{M} \sim \mathrm{Gumbel}(\mathbf{0}, \mathbf{1}),
\end{align}
where $f(\cdot)$ is sigmoid or softmax and noise is sampled from Gumbel distribution \citep{gumbel1954statistical}. The noise encourages the router to explore all $M$ slots during training, building informative representations across the full memory. It is removed at inference for deterministic routing. \Cref{tab:router_ablations} shows consistent gains from Gumbel noise across configurations, confirming that slot exploration during training translates to better-utilized memory at test time.

\subsubsection{Linear vs.\ MLP Projection}
\label{sec:r4}

The standard routing projection is a single linear layer $\mW\vx_t$. Following \citet{zaya}, who showed that MLP projections yield richer router representations in MoEs, we ablate an MLP variant:
\begin{equation*}
\label{eq:mlp}
\vm_t = f\!\left(\mW_2\,\mathrm{GELU}(\mW_1\vx_t + \vb_1)\right).
\end{equation*}
The MLP router improves on some benchmarks (notably FDA) but underperforms the linear router on others, at the cost of additional parameters. Given the inconsistent gains, we select the best-performing configuration per \Cref{tab:router_ablations} for the final \ours design.

\subsection{Memory Shape: Head Dimension vs.\ Number of Slots}
\label{subsec:memshape}

For a fixed memory budget, increasing the number of slots $M$ requires reducing the per-slot feature dimension $d$, and vice versa. These two quantities play fundamentally different roles: $d$ controls how richly each slot can represent a token, while $M$ controls how many distinct items the model can store without interference. The total memory per layer scales as $M \times d \times H$, where $H$ is the number of heads.

\begin{table*}[t]
\centering
\small
\caption{
\textbf{Memory Shape and $\text{Top}_K$ Ablations.}
Accuracy vs. \#memory slots (fixed budget) and the $\text{Top}_K$ write budget on $\hgray{\text{gray-highlighted}}$ SWDE/FDA/SQuAD and NIAH-1/2/3 (1K--32K).
More slots trade per-slot capacity (\texttt{head\_dim}) for token-wise storage.
Larger $\text{Top}_K$ improves long-context retrieval; when $\text{Top}_K=K$, all slots are written with router-scaled intensity.
}
\resizebox{\textwidth}{!}{
\begin{tabular}{ll | >{\columncolor{gray!10}}c>{\columncolor{gray!10}}c>{\columncolor{gray!10}}c |
cccccc | cccccc | cccccc }
\toprule
\textbf{Mem Slot} & \textbf{Top$_K$} &
\textbf{SWDE} & \textbf{FDA} & \textbf{SQuAD} &
\multicolumn{6}{c|}{\textbf{NIAH-1}}  &
\multicolumn{6}{c|}{\textbf{NIAH-2}}  &
\multicolumn{6}{c}{\textbf{NIAH-3}} \\
\# Num. & &
acc$\uparrow$ & acc$\uparrow$ & acc$\uparrow$ &
1K & 2K & 4K & 8K & 16K & 32K &
1K & 2K & 4K & 8K & 16K & 32K &
1K & 2K & 4K & 8K & 16K & 32K \\
\midrule
128 & 16 & 34.5 & 21.8 & \textbf{38.7} &
99.2 & 99.0 & 98.6 & 99.0 & 97.8 & \textbf{98.8} &
98.6 & 99.0 & 87.6 & 8.0 & 1.0 & 2.0 &
66.0 & 44.4 & 13.6 & 0.0 & 0.0 & 0.0 \\

128 & 32 & \textbf{35.6} & 16.2 & \underline{38.5} &
99.6 & \textbf{100} & \textbf{100} & \textbf{100} & \textbf{99.4} & 98.2 &
99.6 & \textbf{99.8} & 92.6 & 41.8 & 11.8 & 8.6 &
78.4 & 42.0 & 2.4 & 3.4 & 0.4 & \textbf{2.8} \\

128 & 64 & \underline{34.8} & 19.4 & 36.0 &
\textbf{99.8} & 99.6 & 99.0 & 98.4 & 96.0 & 73.6 &
99.2 & 97.0 & 93.0 & 42.4 & 6.0 & 7.0 &
\textbf{80.8} & \textbf{54.4} & 4.4 & 2.8 & \textbf{0.8} & 0.6 \\
\hline
256 & 32 & 31.5 & 19.6 & 36.6 &
99.0 & 98.8 & 99.0 & 99.2 & 98.8 & 96.8 &
98.4 & 98.4 & 95.0 & 18.4 & 1.6 & 2.4 &
69.0 & 40.4 & 11.8 & 1.2 & 0.0 & 0.0 \\

256 & 64 & 32.0 & \textbf{22.9} & 33.3 &
98.2 & 98.4 & 96.8 & 97.4 & 97.8 & 96.8 &
97.2 & 98.0 & 95.4 & 21.0 & 3.6 & 1.8 &
70.0 & 47.8 & 6.0 & 0.8 & 0.0 & 0.0 \\

\rowcolor{rmmblue}
256 & 128 & 34.1 & \underline{22.7} & 35.4 &
\textbf{99.8} & \textbf{100} & 99.4 & 99.8 & \textbf{99.4} & 91.4 &
98.8 & 98.0 & \textbf{98.8} & \textbf{81.6} & \textbf{23.0} & \textbf{8.8} &
76.8 & 43.6 & 13.4 & 1.0 & 0.2 & 0.0 \\

256 & 256 & 33.4 & 19.6 & 32.7 &
99.2 & 99.8 & 98.4 & 97.4 & 97.8 & 81.0 &
\textbf{99.8} & 98.6 & 94.6 & 47.8 & 3.8 & 3.6 &
53.6 & 9.4 & 11.8 & 3.6 & 0.6 & 1.0 \\
\hline
512 & 64 & 28.4 & 14.9 & 7.5 &
99.2 & 99.0 & 97.4 & 97.0 & 90.4 & 65.4 &
99.2 & 99.2 & 73.8 & 18.2 & 2.6 & 2.4 &
56.8 & 40.2 & 10.8 & 1.0 & 0.0 & 0.0 \\

512 & 256 & 29.9 & 15.1 & 3.8 &
94.6 & 92.6 & 87.2 & 79.0 & 67.6 & 34.6 &
95.6 & 84.8 & 3.8 & 1.4 & 0.0 & 2.0 &
30.0 & 4.2 & 0.0 & 0.0 & 0.0 & 0.0 \\

512 & 512 & 34.5 & 22.0 & 33.9 &
99.6 & \textbf{100} & 98.8 & 92.2 & 63.8 & 77.2 &
99.4 & 99.2 & 90.6 & 50.2 & 3.8 & 4.6 &
75.0 & 52.8 & \textbf{20.6} & \textbf{6.8} & 0.2 & 0.6 \\
\bottomrule
\end{tabular}
}
\label{tab:memshape}
\end{table*}

\Cref{tab:memshape} sweeps over budget-matched $(M, d)$ configurations and consistently shows that allocating capacity toward more slots improves recall, even as $d$ shrinks. This suggests that slot granularity is the primary driver of retrieval quality: a finer-grained router has more targets to route to, which reduces the probability that a retrieval-critical token shares a slot with unrelated content and gets overwritten. Per-slot capacity, by contrast, appears to matter less; a smaller $d$ is a mild cost compared to the benefit of increased slot count. We fix $M = 256$ as the default, as it is also consistent with previous linear models at the 400M and 800M scales, such as GLA, GDN, and GSA. \Cref{tab:memshape} further shows that the strong recall performance and length generalization of \ours \textit{hold consistently across a \textbf{wide range of memory shapes}}. In particular, most memory configurations outperform linear baselines such as GDN and Mamba-2 on NIAH and other recall-intensive tasks.

\subsection{Top-K Sparsity}
\label{subsec:topk}

The sparsity parameter $K$ controls how many slots are updated at each step. Even without hard Top-$K$ selection, \ours's sigmoid router already produces naturally sparse routing weights: most slot scores fall close to zero, so the effective number of meaningfully written slots is small. Top-$K$ selection takes this further by completely zeroing out all but the $K$ highest-scoring slots, ensuring that unselected slots are exactly frozen rather than receiving negligible but nonzero updates.

We vary $K$ across all memory shape configurations in \Cref{tab:memshape}. Smaller $K$ enforces stricter selectivity, reducing cross-token interference and improving recall on SWDE. Larger $K$ transitions \ours toward dense updates, progressively recovering SSM-like behavior where every slot is touched at every step. We find that $K = 32$ (out of $M = 256$ slots, i.e., $12.5\%$ occupancy) strikes the best balance: it preserves long-lived traces in lightly-used slots while giving the router enough flexibility to distribute general content across multiple slots when appropriate.

\section{Analysis: Learned Memory Allocation}
\label{sec:memory_allocation_analysis}

A key question for any routed memory model is whether the routing is truly meaningful --- does the model learn to allocate memory purposefully, or does it route arbitrarily? This section provides evidence that \ours develops structured, content-aware memory allocation. We study (i) the distribution of effective sequence lengths across slots, which reveals whether slots specialize in terms of how much they process, and (ii) routing patterns on a retrieval task, which reveals \textit{what} is being stored and where.

\begin{figure}[t]
    \centering
    \includegraphics[width=1\linewidth]{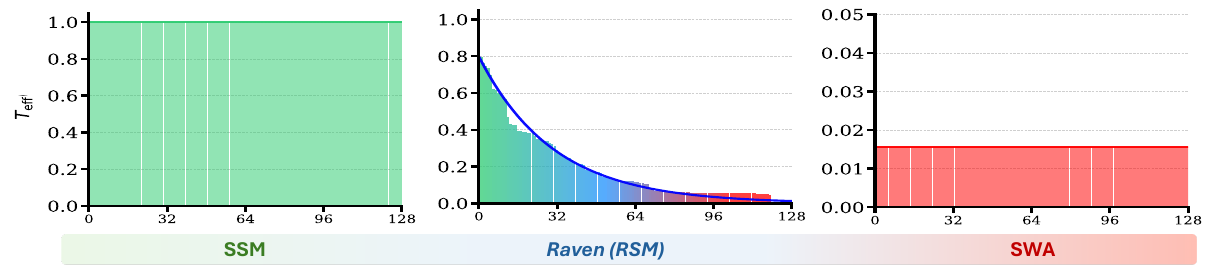}
    \caption{\textbf{\ours Effective Sequence Length.}
     Normalized effective sequence length for a NIAH-1 sample at sequence length 16K.
    $\hred{\textcolor{red!80!black}{\text{SWA}}}$: each token stored in exactly one slot (FIFO).
    $\hblue{\textcolor{blue!80!black}{\text{\ours}}}$: hidden state $\mS_t$ for layer 1, head 1 (256 slots, Top$_{32}$).
    $\hgreen{\textcolor{green!10!black}{\text{SSM}}}$: each token stored in all slots with decay.
    Slots are reordered by usage frequency; results are for 400M parameter models.}
    \label{fig:eslll}
\end{figure}

\subsection{Effective Sequence Length in Routed Memory}
\label{subsec:esl}

The routing mechanism in \ours has a direct consequence on how much each slot processes: a slot that is rarely selected accumulates fewer tokens than one that is frequently written to. To make this precise, we unroll the key recurrence:
\begin{equation*}
\mS^k_t \overset{\text{Unroll}}{=}
\sum_{j=0}^t \prod_{i=j+1}^t \exp(a_i \vr_i) \odot \bigl(\mathbf{1} - \exp(a_j \vr_j)\bigr) \vk_j^\top.
\end{equation*}
The same holds for $\mS^v_t$. Writing $\mS_t = [\mS_t[1], \ldots, \mS_t[M]]$ generically for either state, the update for slot $i$ is:
\begin{equation*}
\mS_t[i] = \sum_{j=0}^t \vk_j \left(\mathbf{1} - \exp(a_j r_j[i])\right) \odot \exp\!\left(\sum_{\ell=j+1}^t a_\ell\, r_\ell[i]\right),
\end{equation*}
where $r_t[i]$ is the $i$-th entry of $\vr_t$. Since $r_t[i] \neq 0$ only when slot $i$ is selected, $\mS_t[i]$ depends only on tokens routed to it. This motivates a \textit{per-slot notion of context length}.

\begin{definition}[Effective Sequence Length]
\label{def:esl}
Given routing weights for slot $i$ across a sequence of length $T$, the \emph{effective sequence length} is:
\begin{equation}
\label{eq:esl}
T_{\mathrm{Eff}}(i) \;=\; \|\vr_i\|_0 \;=\; \sum_{t=1}^{T} \mathbf{1}\!\left[r_{[i,t]} \neq 0\right].
\end{equation}
\end{definition}

We compute the ESL per memory slot and compare \ours to SWA and SSMs in \Cref{fig:eslll}. In SWA, ESL is uniform by construction, each slot receives exactly one token per cycle. In SSMs, ESL is also uniform but maximal, every slot sees every token. \ours breaks this symmetry, with slots falling into three regimes depending on how frequently the router selects them: short ESL ($\hred{\textcolor{red!80!black}{red}}$), long ESL ($\hgreen{\textcolor{green!10!black}{green}}$), and intermediate ($\hblue{\textcolor{blue!10!black}{blue}}$). \Cref{fig:allheadhistmem} extends this analysis across all layers and heads, showing a consistent ESL spread throughout the model.

An important consequence of non-uniform ESL is \textbf{length generalization}. SSMs trained at a fixed context length can degrade on longer sequences~\citep{chen2024stuffed}. State passing addresses this by varying effective lengths during training~\citep{ruiz2025understanding}. \ours achieves the same effect implicitly: within every training example, different slots see different subsequence lengths, naturally exposing the model to variable-length dynamics without any explicit intervention.

\subsection{Selective Allocation in a Retrieval Task}
\label{subsec:retrieval_allocation}

\begin{figure}[t]
    \centering
    \includegraphics[width=1\linewidth]{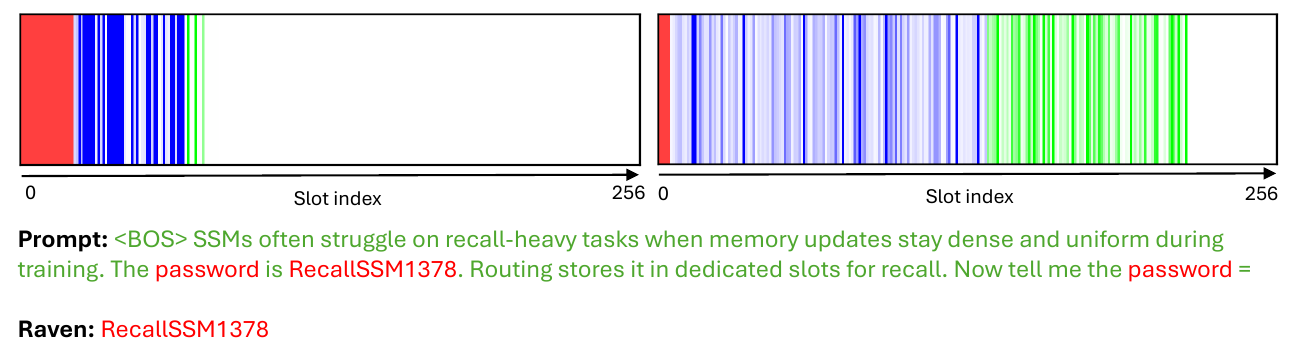}
    \caption{\textbf{\ours Memory Dynamics.}
    Memory allocation for two different heads of \ours on a synthetic NIAH-style task.
    \textcolor{red}{Red} slots store tokens that are important for retrieval (e.g., passwords),
    \textcolor{green!50!black}{Green} slots store non-retrieval tokens, and
    \textcolor{blue}{Blue} slots are shared memory slots between the two type of tokens (red and green). Different heads allocate different amounts of slots to retrieval-important tokens, showing non-uniform memory allocation across heads.}
    \label{fig:vismem}
\end{figure}

Non-uniform ESL alone does not tell us \textit{what} gets routed where. To answer this, we visualize slot assignments on a synthetic retrieval task (\Cref{fig:vismem}). The picture is clear: some slots specialize as \emph{retrieval slots}, receiving almost exclusively passkey tokens~\citep{mohtashami2023landmark} (\textcolor{red}{red}), while others are reserved for non-retrieval content (\textcolor{green!50!black}{green}). Shared slots (\textcolor{blue}{blue}) are rare in heads with strong retrieval behavior. These retrieval slots are analogous to retrieval heads~\citep{bick2504understanding} in Transformers, but operate at finer resolution over the continuous state $\mS_t \in \R^{M \times d}$. By routing retrieval-critical tokens to dedicated slots, \ours limits destructive overwrite and preserves them across long contexts. \Cref{empirical} shows this translates to high retrieval accuracy up to $\mathbf{16\times}$ the training length.

\subsection{Intentional Memory Imbalance}
\label{subsec:load_balancing}

The uneven allocation above is not a side effect; it is by design. MoE routers typically include a load-balancing loss~\citep{shazeer2017outrageously} to encourage uniform token-to-expert assignment and prevent expert collapse. In our setting, uniform memory usage would be counterproductive: forcing equal ESL across slots would prevent the emergence of retrieval slots and expose stored tokens to unnecessary overwrite.

\ours therefore omits any \textit{load-balancing} auxiliary loss, allowing the router to specialize freely. However, as discussed in \Cref{sec:ablations}, \ours applies Gumbel noise to encourage full exploration of its hidden-state memory and prevent collapse to only a few memory slots. The result, shown in \Cref{fig:phoenix_memories}, is that \ours naturally routes retrieval-critical tokens to dedicated regions of $\mS_t \in \R^{M \times d}$ without any explicit supervision to do so.

\section{Related Work}

\subsection{Linear Sequence Mixers}

\paragraph{Linear Transformers and SSMs.}
Softmax attention \citep{vaswani2017attention} scales quadratically with sequence length, making it costly for long contexts and motivating a growing body of sub-quadratic alternatives.
Within this line, Mamba \citep{mamba} matches transformer baselines at similar scale, but its sequential selective scan slows training.
Mamba-2 \citep{mamba2} improves training throughput by simplifying the model and introducing State Space Duality (SSD), later scaled in follow-up work \citep{bick2025llamba,waleffe2024,falconmamba}.
On the linear-attention side \citep{trans_rnn}, Gated Linear Attention (GLA) \citep{yang2023gated} focuses on faster implementations through optimized kernels and hardware-aware training.

A shared limitation is that both GLA and Mamba-2 rely on diagonal or scalar decay, which does not mix across dimensions of the hidden state and can restrict performance on tasks requiring richer state dynamics \citep{merrill2024illusion}.
DeltaNet \citep{deltanet} addresses this by adopting a recurrence based on the Delta rule \citep{schlag2021linear}, where the forget gate depends on a row of the previous state, inducing non-diagonal decay and enabling mixing across state elements. This is made practical by an efficient parallel training algorithm \citep{deltanet,yang2024gated} that reformulates the recurrence into a chunkwise form, enabling hardware-efficient parallelization despite the non-diagonal structure of the forget gate.

Unlike prior approaches, \ours follows a different design point.
It retains fast training by using diagonal decay at the level of individual memory elements, while coupling the elements within each row via a shared row-level update.
As a result, \ours does not rewrite the full memory at every step, enabling persistence that is useful for language modeling.

\paragraph{Dual-State as Higher-Order Linear Attention.}
The recurrence in \Cref{eq:swa_recurrence} maintains two coupled memory states, the key and value caches.
Its \emph{state update} is linear (a rank-one write routed by a one-hot vector) while the \emph{readout} applies standard softmax attention over the $M$ cached tokens.
Following \citet{hla}, this falls under \emph{higher-order linear attention}, meaning that multiple coupled memories are updated by a linear recurrence (without requiring a softmax-free attention readout).
Models such as GSA \citep{gsa}, ABC \citep{abc}, and SWA \citep{longformer} are also second-order instances ($N=2$), and the definition extends directly to $N$ coupled memories.

Among these, \ours most closely resembles GSA: both are dual-state linear attention models over keys and values, with a softmax readout and normalized writes that treat memory as a stack of slots.
The key distinction is that \ours \textit{selectively writes to a subset of slots}, leaving the rest unchanged, whereas GSA updates all slots at every timestep. This selective updating is precisely what enables persistent memory across timesteps.

\paragraph{Hybrid Architectures}
Following the success of SSMs in language modeling and the strong retrieval capabilities of Transformers, several studies have explored hybrid architectures that combine the two. This line of work divides broadly into two directions. The first is full pretraining: Samba~\citep{samba} and Jamba~\citep{jamba} interleave Mamba layers with full softmax attention blocks, Qwen3~\citep{yang2025qwen3} integrates GDN with softmax attention and MoE channel mixers, and GDN~\citep{yang2024gated} proposes variants that combine Mamba-2, GDN, and full attention within a unified architecture. 
The second direction distills pretrained Transformers into hybrid SSM variants~\citep{mohawk,paliotta2025thinking,wang2025mamballama}. \citet{bick2026retrieval} take this further by showing that preserving only the Gather-and-Aggregate retrieval heads while converting the rest into SSM layers recovers most of the Transformer's recall capability at a fraction of the cost.

\paragraph{Retrieval in LLMs.} A growing body of work has identified in-context retrieval as a primary differentiator between Transformer and SSM performance \citep{zoology,repeat_after_me,mamba_icl}, with theoretical and empirical evidence that SSMs struggle with associative recall and precise copying \citep{wen2024rnnstransformersyetkey,repeat_after_me}. 
\citet{bick2504understanding} localize this limitation further, attributing the performance gap to a small subset of Gather-and-Aggregate (G\&A) heads responsible for in-context retrieval --- suggesting the gap reflects a specific functional deficit rather than a holistic architectural failure. 
Because linear models compress history into a fixed-size state, they cannot fully reproduce G\&A behavior. Our results suggest that \ours recovers a restricted but meaningful form of this capability within the constraints of fixed-size models, selectively retrieving information that is queried in advance.

\paragraph{Length Generalization in LLMs.}

Length generalization is the ability to preserve accuracy when inference sequences exceed training lengths. In Transformer LLMs, degradation is often tied to positional encodings and attention statistics shifting out of the training regime. ALiBi replaces explicit positional embeddings with distance-dependent attention biases, improving extrapolation from shorter training contexts \citep{press2022train}. For RoPE-based models \citep{su2021roformer}, Position Interpolation rescales position indices to keep relative rotations within the trained range, enabling large context extensions with limited fine-tuning \citep{chen2023position}; YaRN further improves stability and data efficiency for such adaptation \citep{peng2024yarn}. In recurrent and SSM-based models, length failures are attributed to ``unexplored states'': hidden states during long rollouts fall outside the training distribution, and broadening state coverage can substantially recover performance \citep{buitrago2025length}. Complementary work analyzes out-of-distribution state-space dynamics to extend length generalization in Mamba \citep{lu2025mamba}. 

\ours connects these perspectives via sparse memory routing: by updating only a subset of memory slots at each step, it reduces overwrite while implicitly exposing training to a distribution of effective write horizons, which improves long-range recall and length generalization.

\subsection{Evaluation Benchmarks}
\label{bg:benchmarks}

\paragraph{Retrieval-Heavy Tasks.} These benchmarks require locating and utilizing evidence present in the input context rather than relying on parametric knowledge. We further divide them into two subsets:
(i) \textbf{Real-world retrieval} includes SQuAD \citep{sqd}, an extractive question-answering benchmark in which answers are spans within a supporting passage; SWDE \citep{arora2024simple}, a structured web data extraction task over raw HTML pages with labeled attributes; and FDA \citep{arora2024simple}, which requires extracting a fixed set of attributes from FDA 510(k) report PDFs (up to ${\sim}20$ pages). These tasks demand locating and, in some cases, aggregating evidence from long-form inputs.
(ii) \textbf{Synthetic retrieval} uses \textsc{RULER} \citep{ruler} with the S-NIAH-\{1,2,3\} tasks, which provide controlled settings in which a model must retrieve a key--value association embedded within a long ``haystack.'' S-NIAH-1 uses a passkey-style setup with a repetitive background; S-NIAH-2 embeds the same key--value format within natural-text haystacks (e.g., essays); and S-NIAH-3 increases difficulty by requiring retrieval of longer values (UUIDs) from similar natural-text haystacks.

\paragraph{Language Modeling Tasks.} These benchmarks place comparatively less weight on contextual retrieval, instead primarily testing factual knowledge and commonsense reasoning. We include PIQA \citep{piqa}, Winogrande \citep{winogrande}, HellaSwag \citep{hellaswag}, and ARC (Challenge and Easy) \citep{arc}.

\subsection{Routing and Mixture of Experts (MoE)}

\textbf{Routing:} Token-to-expert routing is a defining component of \textit{Mixture-of-Experts} (\textit{MoE}) models \citep{moe,fedus2022switch}, enabling efficient channel mixers at larger model scales. In MoEs, the router assigns each token to a small subset of \textit{expert} feed-forward networks (FFNs). Therefore, only the selected experts are activated for each token during inference, improving per-token compute efficiency, especially at large model scales (e.g., Mixtral 8$\times$7B \citep{mixtral}).
MoEs have also been used for channel mixing in hybrid architectures. For example, hybrid models such as Kimi Linear \citep{kimilinear} (based on GDN) and Nemotron \citep{nemot} (based on Mamba-2) use MoE layers instead of gated MLPs as channel mixers. Unlike MoEs, which apply routing within their channel mixers, \ours applies routing in its sequence mixer and within its memory $\mS_t$.
Routing has also been used in prior SSMs such as Sparse State Expansion (SSE) \citep{pan2025scaling}, which routes over the key vector $\vk_t$ and sparsely selects its components.

\textbf{Shared Experts:} In recent large language models, such as DeepSeek-V3 \citep{deepseek3} and DeepSeekMoE \citep{deepseekmoe}, not all experts are selected dynamically by the router.
Instead, a subset of feed-forward networks (FFNs) is {always activated} and processes \emph{all} tokens; these are referred to as \emph{shared experts}. This strategy helps avoid under-training experts in MoEs.

\textbf{Mixture of Memories (MoM).} Mixture of Memories (\textit{MoM}) \citep{mom} applies routing among heads of linear transformers, particularly GDN \citep{yang2024gated}. Therefore, not all tokens are processed by each head; instead, only selected heads store a token in their hidden states. In the spirit of our work, \emph{MoM} can be considered a binary router \( r_t^h \) for each hidden state \( \mS_t^h \), which decides whether token \( t \) is stored in the hidden state of the \( h \)-th head \( \mS_t^h \). Moreover, MoM also includes a shared memory, which is updated for all tokens analogous to shared experts in MoEs.

While MoEs apply routing in the channel mixer and MoM applies routing across the heads of the sequence mixer, \ours applies RSM routing within each head's memory allocation, introducing a \textit{\textbf{new dimension} for sparsity.}

\textbf{Mixture of Heads (MoH).} Mixture of Heads (MoH) applies routing to the attention heads of a softmax Transformer~\citep{moh}. Instead of having all heads process every token in multi-head attention (MHA), MoH routes each token to a subset of selected heads, while a set of shared heads always processes all tokens. This is conceptually similar to MoM, which applies routing across different linear Transformer heads: both route at the granularity of heads, whereas \ours applies routing at a higher resolution within its memory $\mS_t$. Routing over heads could also be combined with the \ours router as a promising direction for future work, enabling extremely sparse writes in the sequence mixer.

\newpage

\section{Conclusion}

We introduced Routing Slot Memories (RSMs), a unified framework that formalizes memory slot selection across linear sequence models, and used it to expose a fundamental gap in existing architectures: no prior model combines sparse, input-dependent routing with explicit decay. \ours fills this gap by decoupling \textit{where} information is written from \textit{how long} it persists, giving the model fine-grained control over memory allocation.

\ours learns to isolate retrieval-critical tokens in dedicated slots and naturally induces variable effective sequence lengths across memory. Relative to SWA, it improves two key aspects: content-dependent memory allocation through routing, and selective decay instead of position-based eviction. Relative to dense SSMs, it writes sparsely and selectively, leaving unselected slots unchanged rather than updating the entire state at every step. This allows dedicated retrieval slots to persist with less interference.

These properties let \ours extrapolate to contexts more than $16\times$ longer than those seen during training, without relying on additional components such as short-range convolutions. On recall-intensive benchmarks, \ours consistently outperforms dense SSMs while remaining competitive on general language modeling, and these gains carry over to hybrid architectures paired with both full attention and SWA.

More broadly, the RSM framework suggests that routing and forgetting are orthogonal design axes that existing models have largely conflated. We hope this perspective opens new directions for sequence models that treat memory allocation as a first-class design choice rather than a byproduct of decay.

\bibliographystyle{plainnat}
\bibliography{biblio}

\newpage
\appendix
\section{Appendix}

\subsection{Notation}
\begin{table}[H]
\caption{\textbf{Notations.} Summary of notations used throughout the paper.}
    \centering
    \label{tab:notation}
    \begin{tabular}{ll}
        \noalign{\hrule height 1.5pt}
         \textbf{Definition}& \textbf{Notation} \\
         \hline
$\displaystyle \mS_t$ & Matrix-valued Hidden state \\
$\displaystyle \vq, \vk, \vv$ & Query, Key, Value vectors\\
$\displaystyle \mQ, \mK, \mV$ & Query, Key, Value matrices\\
$\displaystyle a_t$ & Scalar decay \\
$\displaystyle \mA_t$ & Matrix decay \\
$\displaystyle \vr_t$ & Router \\
$\displaystyle \sigma(.)$ & Sigmoid function \\
$\displaystyle \phi(.)$ & Non-linear function \\ 
$\displaystyle \odot$ & Hadamard product\\ 
\noalign{\hrule height 1.5pt}
\end{tabular}
\end{table}

\subsection{Slot Separability of Linear Updates} \label{slotsepdt}

\paragraph{Proposition.}
Let $\mS_{t-1}\in\R^{M\times d}$, $\mD_t\in\R^{M\times M}$,
$\mA_t\in\R^{d\times d}$, and $\mU_t\in\R^{M\times d}$, and consider
\begin{equation}
\mS_t \;=\; \mD_t \mS_{t-1} \mA_t \;+\; \mU_t.
\label{eq:affine-update}
\end{equation}
Denote by $\mS_t[i] \in \R^{1\times d}$ the $i$-th row of $\mS_t$.
Then the map $\mS_{t-1}\mapsto \mS_t$ is \emph{row-wise} 
(i.e., $\mS_t[i]$ depends only on $\mS_{t-1}[i]$) 
if and only if $\mD_t$ is diagonal.

\paragraph{Proof.}

($\Leftarrow$) If $\mD_t=\mathrm{diag}(d_1,\dots,d_M)$, then for each $i$,
\[
\mS_t[i]
=
d_i\, \mS_{t-1}[i] \mA_t
+
\mU_t[i],
\]
which depends only on $\mS_{t-1}[i]$. Hence the update is row-wise and hence slot-separable.

($\Rightarrow$)
Assume that the update is row-separable and $\mD_t$ has a nonzero off-diagonal entry $\mD_t[ik] \neq 0$ for some $k \neq i$.
Choose $\Delta \in \R^{1 \times d}$ such that $\Delta \mA_t \neq \mathbf{0}$, which is possible because $\mA_t\ne0$.
Now consider two states $\mS_{t-1}$ and $\mS'_{t-1}$ that differ only in row $k$, with
\[
\mS'_{t-1}[\ell]
=
\begin{cases}
\mS_{t-1}[k] + \Delta, & \ell = k,\\
\mS_{t-1}[\ell], & \ell \neq k.
\end{cases}
\]
Then,
\[
\mS'_t[i] - \mS_t[i]
=
\mD_t[ik]\,\Delta \mA_t
\neq \mathbf{0}.
\]
Thus, modifying only row $k \neq i$ changes $\mS_t[i]$, contradicting row-separability.
Therefore all off-diagonal entries of $\mD_t$ must be zero, and hence for slot-separable update $\mD_t$ must be diagonal.
\qed

\subsection{Experimental Details}
\label{app:experimental_details}

\subsubsection{Training Recipe}

Our experiments follow the standard pipeline in the literature \citep{yang2023gated,yang2024gated,deltanet,mamba2}, specifically that of \cite{deltanet}. Models are trained at two scales: $\sim$400M and $\sim$800M parameters. All models are trained on SlimPajama-627B \citep{simpij}. The 400M models are trained for 15B tokens with a sequence length of 2048, while the 800M models are trained for 32B tokens with a sequence length of 4096. We use a batch size of 0.5M and the Adam optimizer \citep{kingma2014adam} with a learning rate of $4 \times 10^{-4}$ \citep{yang2024gated}. The learning rate follows a cosine scheduler with 1B tokens of warmup. We utilize the  \texttt{flash-linear-attention}\footnote{\href{https://github.com/fla-org/flash-linear-attention}{https://github.com/fla-org/flash-linear-attention}}  repository for baselines as well as our codebase.

\subsubsection{Models Configuration}

Our experiments encompass several state-of-the-art linear models, including GLA, GSA, Mamba-2, GDN, and SWA. We follow their original configurations at both experimental scales. Importantly, to ensure a fair comparison, particularly on retrieval tasks, \textbf{we carefully match the overall memory elements} across all models. Since single-state SSMs such as GLA and GDN employ only a single recurrence in their original configurations \citep{yang2024gated}, we expand the value dimension of these models so that they share the same number of memory elements as other baselines.

All models use 24 layers, except for Mamba-2, which uses 48 layers as it does not rely on a channel mixer. For GLA, we include the short-range convolution, as it has been shown to significantly improve performance \citep{deltanet}; therefore, all our GLA baselines use short-range convolution. We also use the specialized Triton kernels provided by the \texttt{flash-linear-attention} repository for each model.

Transformer baselines all use 24 layers and 16 heads, following the LLaMA architecture \citep{dubey2024llama}. We include three different positional encoding schemes, RoPE \citep{su2021roformer}, NoPE \citep{nope}, and the Forgetting Transformer (FoX) \citep{fox}, to ensure comprehensive evaluation. The number of memory slots for dual-state recurrences is set to 256. 

\subsection{\ours Recurrence}  \label{phoenix_fullrec}

\begin{tcolorbox}[colback=rmmblue!30!white, colframe=black!100!white, boxrule=0.3mm, arc=2mm, boxsep=0.1pt, left=2mm, right=4pt, top=4pt, bottom=4pt]
\textbf{\ours Recurrence:}
\begin{align}
\notag \vm_t &= \sigma(\mW\vx_t), \quad \vg_t = 
    \begin{cases}
        \vm_{t}[i] &  i \in \text{Top}_K(\vm_{t}) \\
       0.0& \textit{o.w}
    \end{cases},
  \vr_t= \frac{\vg_{t}}{\alpha\sum_{i=1}^{M} \vg_{t}[i]}, \quad a_t = -\text{SoftPlus}(\vw^\top\vx_t) . \exp(\Delta), \\
   \notag \mS^{k}_t &= \exp(a_t \vr_t)  \odot \mS^{k}_{t-1}  +  (\mathbf{1} - \exp(a_t \vr_t))\vk_t^\top, \quad
\mS^{v}_t = \exp(a_t \vr_t)  \odot  \mS^{v}_{t-1}   + (\mathbf{1} - \exp(a_t \vr_t))\vv_t^\top .\\
   \notag \vo_t &= (\mS^{v}_t)^\top \,\softmax\!\left(\mS^{k}_t  \vq_t\right).
\end{align}
\end{tcolorbox}

\subsection{Efficient Training using Gated Linear Attention} \label{sec:trainhow}

As \ours recurrence for keys and values, both can be viewed as Gated Linear Attention (GLA) as presented in \citet{yang2023gated}. We first briefly describe the chunkwise parallel training strategy used in GLA models, and then present two efficient algorithms for training \ours based on GLA kernels.

GLA applies a linear recurrence of the form:
\begin{align*}
    \vo_t = \text{GLA}\{\vq_t,\vk_t,\vv_t,\mG_t\} : \mS_t &= \mS_{t-1} \odot \mG_t + \vv_t\vk_t^\top, \quad \vo_t = \mS_t\vq_t,
\end{align*}
where $\mG_t \in \mathbb{R}^{M\times d}$ is a two-dimensional forget gate. Assuming a diagonal forget gate $\mA_t$, we can rewrite GLA as:
\begin{align*}
    \vo_t = \text{GLA}\{\vq_t,\vk_t,\vv_t,\mG_t\} : \mS_t &= \mS_{t-1} \mA_t + \vv_t\vk_t^\top, \quad \vo_t = \mS_t\vq_t.
\end{align*}
By unrolling the above recurrence and only focusing in the output, we obtain:
\begin{align*}
    \vo_t = \left( \sum_{\tau=0}^t \left( \prod_{i=\tau+1}^t \mA_i \right) \vv_\tau \vk_\tau^\top \right) \vq_t
    = \left( \sum_{\tau=0}^t \left( \prod_{i=0}^t \mA_i \prod_{j=0}^\tau \mA_j^{-1} \right) \vv_\tau \vk_\tau^\top \right) \vq_t.
\end{align*}
We can incorporate the cumulative forget gates into the query/key pairs, leading to:
\begin{align*}
    \vo_t = \sum_{\tau=0}^t \vv_\tau \left( \prod_{j=0}^\tau \mA_j^{-1} \vk_\tau \right)^\top \left( \prod_{i=0}^t \mA_i \vq_t \right).
\end{align*}
As $\mA_t = \mathrm{diag}(\va_t)$, we define the cumulative product of these decay vectors as
\[
\vb_t = \prod_{i=\tau+1}^{t} \va_i,
\]
and stack them over a sequence of $T$ tokens into a matrix $\mB \in \mathbb{R}^{T \times d}$. Similarly by shaping the query, key, and value matrices, we can express the parallel form of the output $\mO \in \mathbb{R}^{T \times d}$ as \cite{yang2023gated}:
\begin{align}
\label{eq:pararnn}
    \mO = \left(\left( \left(\mQ \odot \mB \right)\left(\frac{\mK}{\mB} \right)^\top \right) \odot \mM \right) \mV,
\end{align}
where $\mM \in \mathbb{R}^{T \times T}$ is a binary causal mask. Since \Cref{eq:pararnn} may suffer from numerical instability for large sequence lengths $T$, due to the cumulative product of small values in $\mB$, parallel training applies \Cref{eq:pararnn} in a chunkwise manner. Specifically, the computation is performed over $n$ chunks of sequence length $C$, with $T = n \times C$. This algorithm results in efficient and fast training for GLA models using specialized GPU kernels. 

Considering GLA's efficient training presented as $\text{GLA}(\vq_t,\vk_t,\vv_t,\mG_t)$ we now present the efficient algorithms for training \ours.

Looking at \ours recurrence presented in \Cref{phoenix_fullrec} we can re-write the output of \ours recurrence ($\vo_t$) as two-pass GLA as used in GSA \citep{gsa}, followed by:
\begin{tcolorbox}[
  colback=gray!5!white,
  colframe=gray!5!white, boxrule=0pt,           arc=0mm,               boxsep=0.1pt,
  left=2mm, right=4pt, top=4pt, bottom=4pt
]
\begin{align*}
\textbf{RSM Parallel Form:} \\
    \vo'_t &= \text{GLA}\{\vq_t,\vk_t,1-\exp(a_t\vr_t),\exp(a_t\vr_t)\} \\
    \vo_t  &= \text{GLA}\{\softmax(\vo'_t),\vv_t,1-\exp(a_t\vr_t),\exp(a_t\vr_t)\}
\end{align*}
\end{tcolorbox}
Therefore, we can directly adapt efficient implementation of GSA kernels using flash-linear attention (FLA) library \citep{fla}. We can use two separate passes of GLA for each hidden states and their parallel forms respectively and measure the output as shown above.

\subsection{\ours Design Details}
In this section, we describe in detail all our design choices for \ours and final block design of \ours sequence mixer and channel mixer in both SSM and hybrid architectures.

\paragraph{\ours Linear Transformer} \ours follows the standard 400M-parameter model design used in the SSM literature \citep{yang2023gated,deltanet,yang2024gated,movahedi2025selective}. We use four heads per block and separate per-head hidden states $\mS^v_t$ and $\mS^k_t$ for values and keys. We also use a channel mixer after each RSM block, with skip connections \citep{resnet} as in Transformers and SSMs. We use QK normalization and apply output gating in the RSM block. We use the Llama tokenizer \citep{dubey2024llama} for \ours and all other models.

\paragraph{\ours Hybrid Model}  In the hybrid setup we interleaved \ours RSM block with SWA and softmax attention mixers. After each sequence mixer, gated MLP was used as the channel mixer with skip connections similar to hybrid models setup of \citet{yang2024gated}. Softmax attention and SWA mixer of hybrid \ours use NoPE as position embedding. We have also evaluated RoPE as position embedding for SWA and attention mixers of hybrid \ours and observed their failure in length generalization (shown in \Cref{tab:full_hybrid_comparison}).

In the SWA hybrid setting, greater pressure is placed on the linear component, since SWA layers can only retrieve information within a fixed local window and cannot directly attend to arbitrary positions. 
Importantly, SWA-RoPE does not harm length generalization in this setting, because every query–key pair is bounded by the window size; thus RoPE never encounters relative distances outside its training distribution regardless of the total sequence length.
As a result, long-range retrieval falls entirely on the linear layer. 
\ours+SWA with RoPE achieves near-perfect NIAH-1 accuracy across the full 1K--32K range, while GDN+SWA degrades steadily and Mamba-2+SWA collapses to near zero beyond 4K. 
In this configuration \ours also leads on SWDE, FDA, and SQuAD, confirming that its advantages are architectural rather than dependent on a particular hybrid pairing.
\paragraph{Initialization.}
As \ours readout uses the softmax function,
$\vo_t = (\mS^{v}_t)^\top \,\softmax\!\left(\mS^{k}_t \vq_t\right)$,
and the memory slots are initialized to zero, empty slots contribute a value of 1 in the readout (since $\exp(0)=1$ in the softmax).
In our initial experiments, we avoided this effect by dynamically masking empty slots with $-\infty$, as is done in softmax causal attention.
However, we observed no difference when omitting this mask and using zero-initialized states directly in the readout.
Therefore, we adopt this simpler setting for \ours.

\begin{table*}[t]
\centering
\caption{
\textbf{Hybrid Models Retrieval Ability.} The performance of state-of-the-art linear models used as hybrid models with different attention types. Performance is reported on NIAH-1,2,3 and $\hgray{\text{{Gray-highlighted}}}$ SWDE, FDA, and SQuAD.
}
\resizebox{\textwidth}{!}{
\begin{tabular}{lllc | >{\columncolor{gray!10}}c>{\columncolor{gray!10}}c>{\columncolor{gray!10}}c |
cccccc | cccccc | cccccc }
\toprule
\textbf{Hybrid} & \textbf{Positional} & \textbf{Linear} & \textbf{No} &
\textbf{SWDE} & \textbf{FDA} & \textbf{SQuAD} &
\multicolumn{6}{c|}{\textbf{NIAH-1}}  &
\multicolumn{6}{c|}{\textbf{NIAH-2}}  &
\multicolumn{6}{c}{\textbf{NIAH-3}} \\
\textbf{Component} & \textbf{Embedding} & \textbf{Model} & \textbf{Conv.} &
acc$\uparrow$ & acc$\uparrow$ & acc$\uparrow$ &
1K & 2K & 4K & 8K & 16K & 32K &
1K & 2K & 4K & 8K & 16K & 32K &
1K & 2K & 4K & 8K & 16K & 32K \\
\midrule
\multirow{3}{*}{Attn} & \multirow{3}{*}{RoPE} & GDN & \textcolor{red}{\xmark}
& 47.1 & 22.3 & 34.2
& \textbf{100} & \textbf{100} & \textbf{53.4} & \textbf{23.4} & \textbf{11.6} & \textbf{6.0}
& \textbf{100} & \textbf{100} & \textbf{53.0} & \textbf{20.2} & \textbf{10.4} & \textbf{5.3}
& 97.2  & 87.8  & \textbf{27.0} & \textbf{9.4} & \textbf{2.4} &0.0\\
&  & Mamba-2 & \textcolor{red}{\xmark}
& \textbf{52.2} & \textbf{26.1} & \textbf{37.7}
& \textbf{100} & 96.8 & 16.8 &0.0&0.0& 0.0
& \textbf{100} & \textbf{100} & 1.6 &0.0&0.0& 0.0
& \textbf{99.4}  & \textbf{91.0}  &0.0&0.0&0.0&0.0\\
\rowcolor{rmmblue}
\cellcolor{white}& \cellcolor{white} & \ours & \greencheck
& 34.5 & 9.7 & 33.3
& \textbf{100} & 77.8 &0.0&0.0&0.0& 0.0
& \textbf{100} & 99.8 &0.0&0.0&0.0& 0.0
& 71.8  & 62.0  &0.0&0.0&0.0&0.0\\
\midrule
\multirow{3}{*}{Attn} & \multirow{3}{*}{NoPE} & GDN & \textcolor{red}{\xmark}
& 54.6 & 67.2 & 34.5
& \textbf{100} & \textbf{100} & \textbf{100} & \textbf{100} & 93.2 & 70.5
& \textbf{100} & \textbf{100} & \textbf{100} & 8.0 &0.0& 0.0
& \textbf{94.2}  & 70.2  & 50.0  &0.0&0.0&0.0\\
&  & Mamba-2 & \textcolor{red}{\xmark}
& \textbf{56.3} & \textbf{68.8} & \textbf{36.0}
& \textbf{100} & \textbf{100} & 16.4 &0.0&0.0& 0.0
& \textbf{100} & \textbf{100} & 85.8 &0.0&0.0& 0.0
& 79.6  & \textbf{80.6}  & 60.8 &0.0&0.0&0.0\\
\rowcolor{rmmblue}
\cellcolor{white} & \cellcolor{white} & \ours & \greencheck
& 51.4 & 64.2 & 31.4
& \textbf{100} & \textbf{100} & \textbf{100} & \textbf{100} & \textbf{98.4} & \textbf{78.6}
& \textbf{100} & \textbf{100} & \textbf{100} & \textbf{100} & \textbf{95.4} & \textbf{65.4}
& 89.6  & 67.0  & \textbf{73.8}  & \textbf{60.0} & \textbf{10.2} & \textbf{14.4} \\
\midrule
\multirow{3}{*}{SWA} & \multirow{3}{*}{RoPE} & GDN & \textcolor{red}{\xmark}
& 30.4 & 14.7 & 33.0
& 78.2 & 75.8 & 58.8 & 54.4 & 45.0 & 22.1
& 68.6 & 46.6 & 11.0 & 3.8 & 3.2 & 1.3
& 55.4 & 23.0 & 8.6 & 2.0 & \textbf{2.0} & 1.1 \\
&  & Mamba-2 & \textcolor{red}{\xmark}
& 12.0 & 9.3 & 36.1
& 29.8 & 11.0 & 6.2 & 3.4 & 1.2 & 0.6
& 33.4 & 16.8 & 5.8 & 5.2 & 3.2 & \textbf{2.8}
& 4.2  & 2.8  & 6.6 & \textbf{2.2} &0.0& 1.4 \\
\rowcolor{rmmblue}
\cellcolor{white}& \cellcolor{white} & \ours & \greencheck
& \textbf{33.0} & \textbf{25.9} & \textbf{37.0}
& \textbf{100} & \textbf{99.6} & \textbf{99.6} & \textbf{99.8} & \textbf{99.2} & \textbf{99.6}
& \textbf{99.6} & \textbf{99.8} & \textbf{95.2} & \textbf{51.0} & \textbf{6.8} & 2.4
& \textbf{64.4}  & \textbf{42.4}  & \textbf{12.0} & 0.8 & 1.8 & \textbf{2.4} \\
\midrule
\multirow{3}{*}{SWA} & \multirow{3}{*}{NoPE} & GDN & \textcolor{red}{\xmark}
& \textbf{35.6} & \textbf{23.9} & 32.4
& \textbf{99.8} & \textbf{100} & \textbf{98.6} & \textbf{59.8} & \textbf{26.8} & 10.2
& \textbf{99.2} & \textbf{96.6} & \textbf{42.0} & 6.6 & \textbf{3.6} & 1.3
& \textbf{77.2} & \textbf{53.8} & \textbf{24.4} & \textbf{2.6} & 1.2 &0.0\\
&  & Mamba-2 & \textcolor{red}{\xmark}
& 13.9 & 14.6 & 32.9
& 30.0 & 11.0 & 6.2 & 0.4 &0.0& 0.0
& 33.4 & 16.8 & 5.4 &0.0&0.0& 0.0
& 18.8 & 7.8  & 4.6 &0.0&0.0&0.0\\
\rowcolor{rmmblue}
\cellcolor{white} & \cellcolor{white} & \ours & \greencheck
& 24.9 & 22.2 & \textbf{33.8}
& 63.2 & 51.4 & 44.4 & 38.8 & 25.2 & \textbf{17.8}
& 54.0 & 40.8 & 16.0 & \textbf{7.8} & \textbf{3.6} & \textbf{2.8}
& 16.6 & 7.0  & 2.0  & 0.6 & \textbf{1.6} & \textbf{1.2} \\
\bottomrule
\end{tabular}
}
\label{tab:full_hybrid_comparison}
\end{table*}

\subsection{Test-Time Training Point of View}

Test-Time Training (TTT)~\citep{ttt} interprets the recurrence in linear transformers as an online learning update driven by a per-token objective. Consider the linear attention recurrence
\begin{align}
    \mS_t = \mS_{t-1} + \vk_t \vv_t^{\top}.
\end{align}
For clarity, we transpose the content matrix $\mU_t$ as $\vk_t \vv_t^\top$ so that it aligns with the TTT framework~\citep{zhang2025test}. This update can be viewed as a single step of gradient descent on the online loss
\begin{align}
    \mathcal{L}(\mS) = -\langle \mS\vk_t,\, \vv_t\rangle,
    \qquad
    \mS_t = \mS_{t-1} - \nabla_{\mS}\mathcal{L}(\mS),
\end{align}
where $\mathcal{L}$ serves as the online learning objective. By varying the choice of online objectives and the corresponding gradient-based update rules, a broad family of linear recurrent models can be recovered. An overview of existing diagonal linear recurrence models under this perspective is provided in \Cref{tab:ttt_overview}.

As shown in \Cref{tab:ttt_overview}, the forget gates used in different linear transformer architectures can be interpreted through a shared duality with weight decay~\citep{krogh1992simple} in the online learning objective $\mathcal{L}$ (corresponding to the $\lVert \mS \rVert$ regularization term). The router in \ours extends this objective by introducing a form of \textit{\textbf{selective Dropout}}~\citep{dropout} into the online learning loss, resulting in sparse state updates. This dropout effect primarily arises from the {\textit{sparsity of the router}} $\vr_t$ and the associated online regularizer $1-\textcolor{blue}{\exp(a_t \vr_t)}$, shown in blue: when slot $i$ is not selected ($\vr_t[i]=0$) the regularizer vanishes, so that slot incurs no weight decay and its content is preserved exactly. One can look at \ours update as applying \textit{\textbf{sparse TTT }}and only updating selected slots.

\begin{table}[t]
\centering
\caption{An overview of different attention mechanisms through the lens of state updating rules and their learning objective under the TTT framework \citep{ttt}. Table style is inspired by \citet{kimilinear}.}
\resizebox{\textwidth}{!}{
\begin{tabular}{@{}l| p{0.52\linewidth} p{0.38\linewidth}@{}}
\toprule
 \textbf{Model} & \textbf{Objective $\mathcal{L}$} & \textbf{Update rule $\mS_t = \mS_{t-1} - \nabla_{\mS}\mathcal{L}(\mS)$} \\
\midrule
{Linear Attention} \citep{trans_rnn}
& $\displaystyle -\langle \mS_{t-1}^{\top}\vk_t,\, \vv_t\rangle$
& $\displaystyle \mS_t = \mS_{t-1} + \vk_t \vv_t^{\top}$ \\
{RetNet} \citep{retnet}
& $\displaystyle -\beta_t\langle \mS_{t-1}^{\top}\vk_t,\, \vv_t\rangle \;+\; \frac{1}{2}\left\lVert \sqrt{{1-a}}\, \mS_{t-1}\right\rVert_F^2$
& $\displaystyle \mS_t = a \mS_{t-1} + \beta_t \vk_t \vv_t^{\top}$ \\
{Mamba-2} \citep{mamba2}
& $\displaystyle -\beta_t\langle \mS_{t-1}^{\top}\vk_t,\, \vv_t\rangle \;+\; \frac{1}{2}\left\lVert \sqrt{{1-a_t}}\, \mS_{t-1}\right\rVert_F^2$
& $\displaystyle \mS_t = a_t \mS_{t-1} + \beta_t \vk_t \vv_t^{\top}$ \\
{GLA} \citep{yang2023gated}
& $\displaystyle -\langle \mS_{t-1}^{\top}\vk_t,\, \vv_t\rangle \;+\; \frac{1}{2}\left\lVert \sqrt{{\mathrm{Diag}(1-a_t)}}\, \mS_{t-1}\right\rVert_F^2$
& $\displaystyle \mS_t = {\mathrm{Diag}(a_t)}\, \mS_{t-1} + \vk_t \vv_t^{\top}$ \\
\hline
\rowcolor{rmmblue}
\textbf{\ours} & $\displaystyle -(1-\textcolor{blue}{\exp(a_t\vr_t)})\langle \mS_{t-1}^{\top}\vk_t,\, \vv_t\rangle \;+\; \frac{1}{2}\left\lVert \sqrt{1-\textcolor{blue}{\exp(a_t\vr_t)}}\, \mS_{t-1}\right\rVert_F^2$
& $\displaystyle \mS_t = \mS_{t-1}\,{\mathrm{Diag}(\textcolor{blue}{\exp(a_t\vr_t)}}) + (1-\textcolor{blue}{\exp(a_t\vr_t)})\,\vk_t \vv_t^{\top}$ \\
\bottomrule
\end{tabular}
}
\label{tab:ttt_overview}
\end{table}

Models using the Delta update rule, such as DeltaNet and KDA, apply a different online learning objective, which takes the form:
\begin{align}
    \mathcal{L}(\mS) = \tfrac{1}{2}\lVert \mS^{\top}\vk_t-\vv_t\rVert^2,
\end{align}
This online update rule has connections to \textit{Associative Memory} introduced in Hopfield Networks \citep{hophop}. \textit{We refer to Table 7 of \citet{kimilinear} for more details on the TTT framework.}

\begin{figure}[h]
    \centering
    \includegraphics[width=1\linewidth]{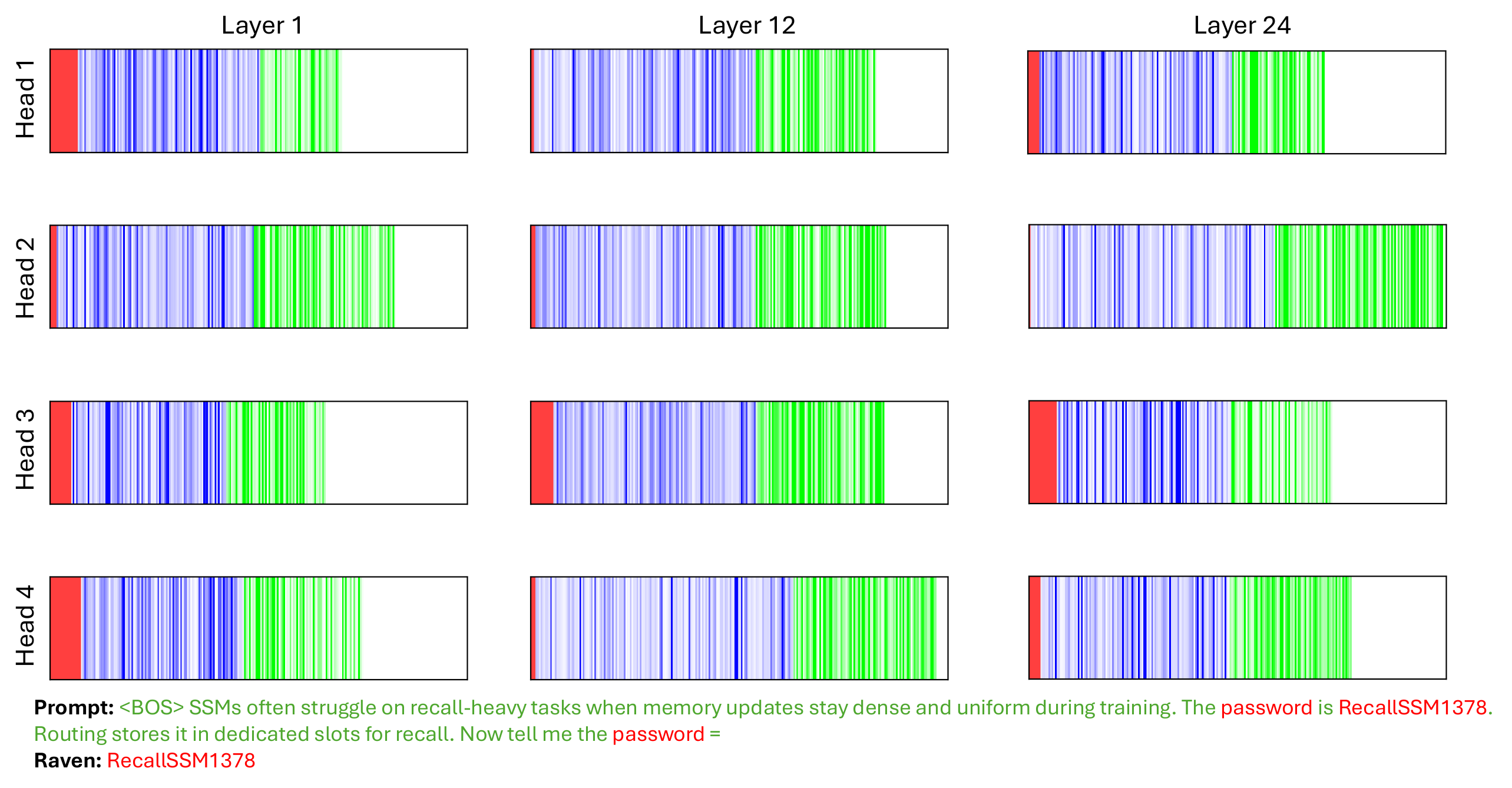}
     \caption{\textbf{Memory Visualization.} Memory of \ours while being asked the password within the prompt. The figures show the hidden state of \ours $\mS_t$ at the end of processing the prompt and answering to the question (time point $t=L$) for \textbf{(Top)} Layer 24, \textbf{(Middle)} Layer 12, \textbf{(Bottom)} Layer 2 for the first head among 4 heads at 400M scale. \textcolor{red}{Red} slots are storing only retrieval (red) tokens, \textcolor{green!50!black}{green} slots are memory slots only storing the normal (green) tokens, and \textcolor{blue}{blue} are shared memory slots between both token types. }
    \label{fig:phoenix_memories}
\end{figure}

\subsection{Memory Dynamic Visualizations}
\label{app:memory_visualizations}

\begin{figure}
    \centering
    \includegraphics[width=1\linewidth]{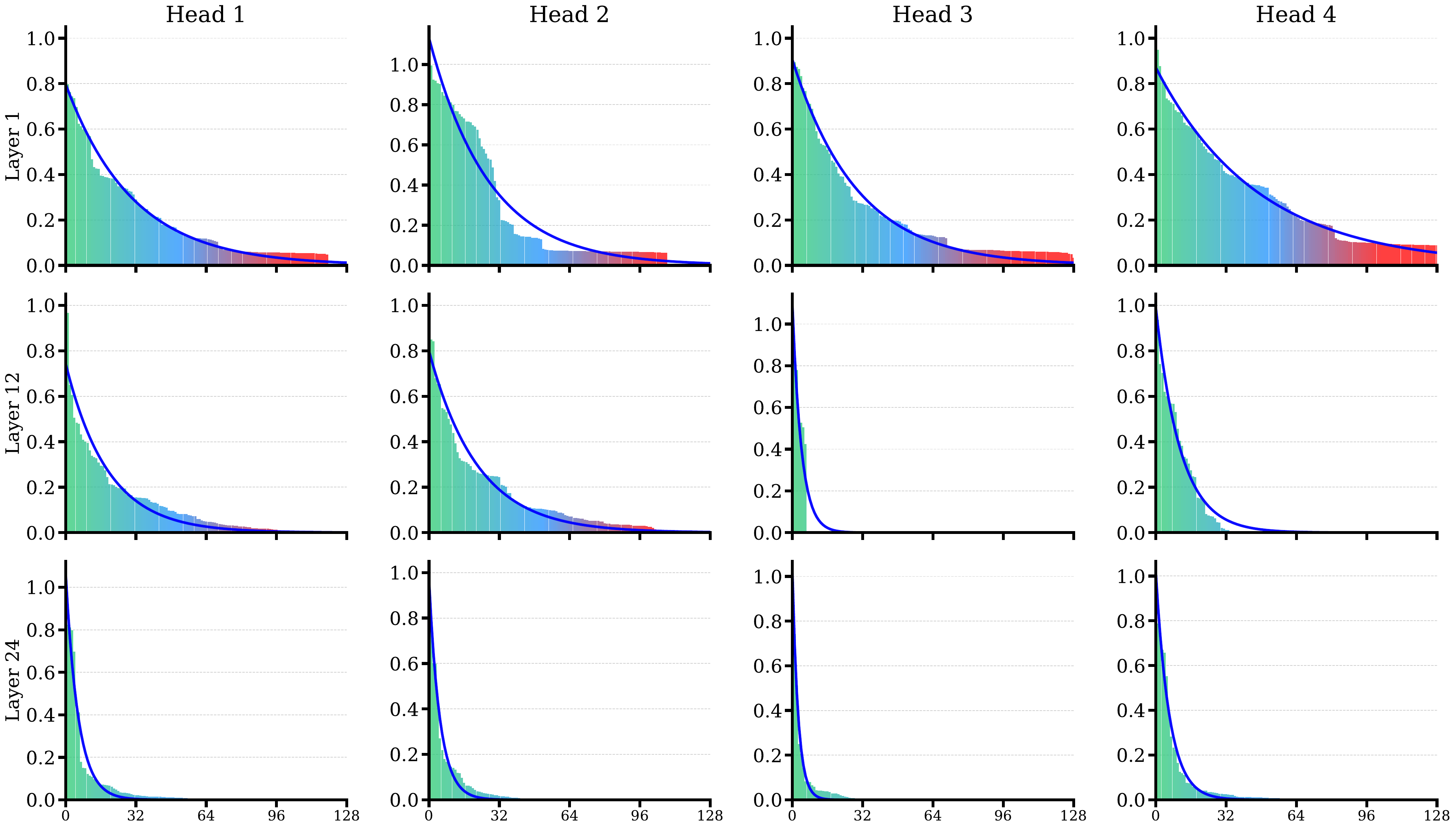}
    \caption{\textbf{Effective Sequence Length of \ours.} The normalized \textit{effective sequence length} for a NIAH-1 sample with a sequence length of 16K. Results are from the first, 12th, and last (24th) layers, for all four heads of the \textit{400M} model. Slots are reordered from highest to lowest.}
    \label{fig:allheadhistmem}
\end{figure}

We visualize the memory dynamics of \ours. As shown in \Cref{fig:allheadhistmem}, different heads in \ours exhibit different Effective Sequence Lengths (\textit{ESL}). Some heads are extremely sparse; for example, head 3 in layer 24 activates only a few memory slots to store tokens. In contrast, other heads, such as head 4 in layer 1, activate many more slots and store tokens in a more diverse manner.

\Cref{fig:phoenix_memories} further illustrates how different memory slots in \ours process the same simple text differently and how tokens are distributed across slots. Specifically, \textcolor{red}{red} regions indicate slots that store passkey tokens, while \textcolor{blue}{blue} regions represent slots shared between passkey and non-passkey tokens. As observed, only a few heads are responsible for retrieving and storing passkey tokens without overwriting them (red regions). In contrast, some heads, such as head 2 in layer 24, do not allocate dedicated memory slots for retrieval tokens; instead, they automatically use shared memory for both retrieval and regular tokens.

\Cref{fig:vismem} also clearly shows that for a given sentence with a passkey \ours retrieval head clearly stores the password in unique sections (slots) of memory without over-writing it, while for other tokens it will store them in shared parts of memory. Moreover, a non-retrieval head stores all tokens equally in the memory.

\end{document}